\documentclass[final,3p,times]{elsarticle}

\usepackage{amssymb}
\usepackage{amsmath}

\newcommand{\ours}[0]{EFGCN}
\usepackage{hyperref}
\usepackage{subcaption}
\usepackage{tikz}
\usepackage{pgfplots}
\usepackage{booktabs}
\usepackage{graphicx}
\usepackage{pifont}
\usepackage{wrapfig}
\pgfplotsset{compat=1.18}

\newif\ifshowchanges
\showchangesfalse    

\newcommand{\changed}[1]{
  \ifshowchanges            
    {\color{red}#1}%
  \else 
    #1%
  \fi
}

\newcommand{\xmark}{\ding{55}}  

\journal{Nuclear Physics B}

\begin{document}

\begin{frontmatter}



\title{Embedded Graph Convolutional Networks for Real-Time Event Data Processing on SoC FPGAs}


\author[agh]{Kamil Jeziorek}
\author[agh]{Piotr Wzorek}
\author[agh]{Krzysztof Blachut}
\author[sorbonne]{Andrea Pinna}
\author[agh]{Tomasz Kryjak}

\affiliation[agh]{organization={
            Embedded Perception and Autonomous Systems Group,
            Department of Automatic Control and Robotics, 
            AGH University of Krakow},
            country={Poland}}

\affiliation[sorbonne]{organization={
            Sorbonne Universite, 
            CNRS, 
            LIP6, 
            F-75005},
            country={Paris}}

\begin{abstract}

The utilisation of event cameras represents an important and swiftly evolving trend aimed at addressing the constraints of traditional video systems. Particularly within the automotive domain, these cameras find significant relevance for their integration into embedded real-time systems due to lower latency and power consumption. One effective approach to ensure the necessary throughput and latency for event processing is through the utilisation of graph convolutional networks (GCNs).
In this study, we introduce a~custom \ours\ (Event-based FPGA-accelerated Graph Convolutional Network) designed with a~series of hardware-aware optimisations tailored for PointNetConv, a~graph convolution designed for point cloud processing. The proposed techniques result in up to 100-fold reduction in model size compared to Asynchronous Event-based GNN (AEGNN), one of the most recent works in the field, with a~relatively small decrease in accuracy (2.9\% for the N-Caltech101 classification task, 2.2\% for the N-Cars classification task), thus following the TinyML trend.
We implemented \ours\ on a~ZCU104 SoC FPGA platform without any \changed{off-chip} external memory resources, achieving a~throughput of 13.3 million events per second (MEPS) and real-time partially asynchronous processing with low latency.
\changed{Across multiple event-based classification benchmarks, our approach achieves competitive accuracy while providing state-of-the-art computational efficiency per event, small model size, and high scalability, customisability and resource efficiency.}
We publish both software and hardware source code in an open repository: \url{https://github.com/vision-agh/gcnn-dvs-fpga}.

\end{abstract}


\end{frontmatter}



\section{Introduction}

Embedded vision systems have become an integral part of many modern technologies, especially in advanced mobile robotics applications (e.g. autonomous vehicles) \cite{Konstantinidis2021automotive}. Vision sensors facilitate object detection and localisation, which are crucial for navigation, obstacle avoidance, path planning, and performing specific tasks like manipulating objects or interacting with the environment \cite{bodenhagen2014adaptable, pham2015towards}.

Typical frame-based cameras capture video sequences in greyscale or colour with specific spatial (e.g. $1280 \times 720$ pixels) and temporal (e.g. 60 frames per second) resolution.
However, they can be challenging to apply in some circumstances because of issues such as motion blur during fast movements, high latency due to low frame rates, and difficulties in operating in uneven lighting conditions.
These issues could be partially addressed by existing high-speed cameras, but these sensors are expensive and typically require good lighting conditions.
Moreover, frame-based cameras generate large amounts of redundant data, which can be inefficient to process in real-time embedded systems.

In contrast, neuromorphic sensors, also known as event cameras or dynamic vision sensors (DVS), offer several advantages that address these limitations \cite{Lichtsteiner2008sensor}.
In a~DVS, each pixel operates independently and asynchronously\footnote{{In the context of DVS, \textit{asynchronicity} is defined as the ability to generate or process events directly at the moment of an intensity change registered by a~given pixel, without maintaining fixed time intervals.}}, detecting only changes in brightness rather than its absolute value and generating so-called \textit{events}.
Each event encodes the location (pixel coordinates), the timestamp (with microsecond precision), and the polarity (indicating increase or decrease in intensity).
This approach results in lower average power consumption, high temporal resolution, and reduced data redundancy, as the changes are only captured in place and time of their occurrence.
Furthermore, since event generation is based on the logarithm of the brightness change, DVS has a~high dynamic range which enables reliable operation in environments with uneven illumination. 

As highlighted above, event cameras have many advantages, especially for applications in dynamic environments for mobile robotics.
However, efficient processing of the resulting asynchronous event stream remains challenging. Computer vision methods developed over the last \changed{70} years for frame cameras are not suitable for sparse spatio-temporal point clouds. Therefore, several approaches to this problem have been proposed. The simplest method is to project the event data onto a~two-dimensional plane in order to create a~pseudo-frame (often called an event-frame), similar to the one obtained from a~traditional camera \cite{afshar2020event}. This allows the almost direct application of classical methods and convolutional neural networks (CNNs) \cite{6981783, perot2020learning}.
However, this approach requires the aggregation of events over a~fixed time window, which introduces information loss (especially high temporal resolution), generation of redundant pixels, and higher latency. Therefore, recent research has focused on processing events in their original sparse form.

{One recent approach is to use graph neural networks (GNNs) through a~generation of a~graph representation. In such a~structure, individual events are represented as vertices, while the edges connecting these vertices correspond to their local relationships. As a~consequence, the data represented as a~sparse cloud of vertices in spatio-temporal space can be processed efficiently with a~graph neural network. In addition, recent research \cite{li2021graph, schaefer2022aegnn, gehrig2022pushing} indicates that such graphs can be generated and updated asynchronously, thus not only reducing the number of operations for a~single event, but also limiting the latency of data processing.}

{However, a~key challenge remains the efficient processing of event data in a~way that enables real-time\footnote{{The precise definition of 'real-time' depends heavily on the application and sensor used. In 'traditional' computer vision this usually means at least 20-25 frames per second (40-50 ms). In this work, our aim is to process the event stream generated by the sensor without any major delays, i.e. in the order of single microseconds.}} realisation of computer vision tasks with low latency, low power consumption, and high throughput.
It is important to emphasise that meeting these requirements is crucial for applications in mobile robotics, especially in case of e.g. autonomous drone racing.
In this field the manoeuvres during the flight have to be performed as fast as possible, which requires extremely low latency thanks to the algorithm and the computing platform.}

{The literature includes research on hardware implementations of event-based vision algorithms, most commonly for embedded GPU platforms \cite{schaefer2022aegnn, gehrig2022pushing}.
Recently, Field-Programmable Gate Array (FPGA) platforms, including system-on-chip (SoC) FPGA devices, have also gained increasing attention \cite{kryjak2024event, yanh_evgnn}, due to their low power consumption, high parallel processing capabilities, and flexibility enabled by hardware description languages.
There is a~growing collaboration between FPGA manufacturers and the DVS industry, driving advancements in event-based vision\footnote{{https://www.prophesee.ai/event-based-metavision-amd-kria-starter-kit/}}.}

{Therefore, in this work we propose an Event-based FPGA-accelerated Graph Convolutional Network -- our strategy for efficient event data processing aimed at real-time and low-power mobile robotics applications. In contrast to prior FPGA GCN accelerators that assume a~fixed graph structure \cite{wang2022bignn, tao2022lw, zhou2022inference}, EFGCN operates on a~graph that is generated and updated asynchronously from the incoming event stream.} We can summarise our main contribution as follows:
\begin{itemize}

\item {We introduce a~set of hardware-aware optimisations for graph convolutional networks processing event data. These optimisations are guided by FPGA resource constraints and informed by extensive ablation studies on multiple event-based classification datasets.}

\item {We present the first hardware implementation of a~3D MaxPool layer designed for event-data processing, which, by leveraging the temporal sparsity of event-based data, enables the deployment of larger network architectures and facilitates higher classification accuracy.}

\item {We present a~custom hardware implementation of PointNetConv layers that exploits deterministic latency and throughput, achieved by using only constant-latency components without \changed{utilisation of any off-chip} external memory. This design effectively leverages the temporal sparsity of event data for efficient and scalable event-driven processing.}

\item {Using these methods, we present one of the first end-to-end hardware implementations of a~GCN for real-time, continuous, sparse, and asynchronous event data processing. The design is evaluated on a~SoC FPGA through both simulation and hardware testing. To the best of our knowledge, this is the first implementation that supports all characteristic GCN layers, including pooling layers.}


\item \changed{Among hardware-implemented methods considered in this work, \textbf{EFGCN} provides the \textbf{most memory-efficient FPGA approach}, requiring up to \textbf{70\% less on-chip memory} than the closest graph-based hardware baseline while achieving higher accuracy on the N-Cars dataset, demonstrating a~competitive design for mobile robotics applications.}

\end{itemize}

\begin{figure}[t!]

\begin{subfigure}[t]{\linewidth}
    \includegraphics[width=\linewidth]{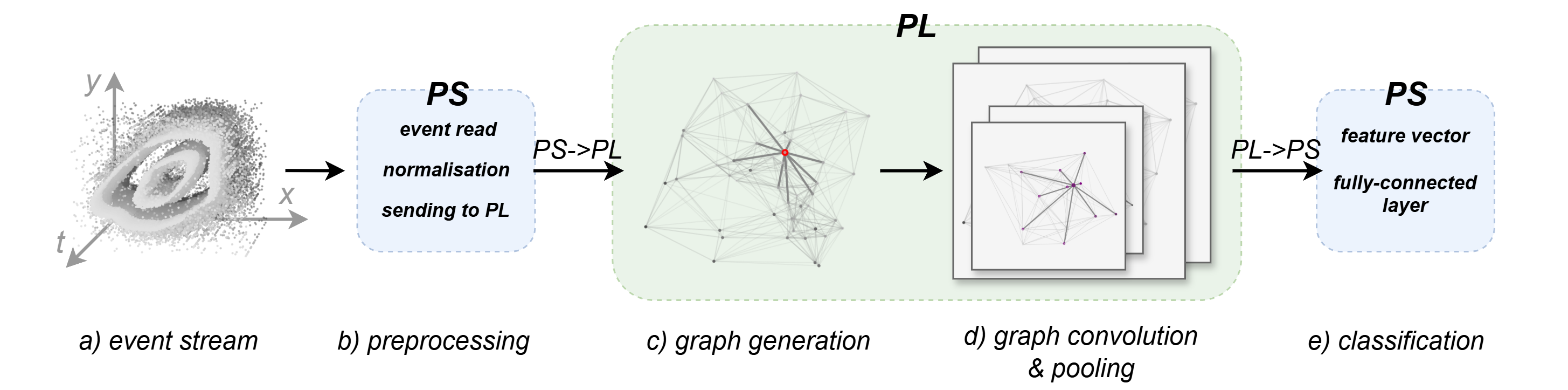}
\end{subfigure}

\begin{subfigure}{\linewidth}
    \begin{tikzpicture}
    \begin{semilogxaxis}[
        tick label style={font=\scriptsize},
        label style={font=\scriptsize},
        legend style={font=\scriptsize},
        width=\linewidth,
        xmin=1000, xmax=100000000,
        height=7cm,
        grid=major,
        xlabel={Number of parameters [\#]},
        ylabel={Accuracy [\%]},
        legend pos= south east,
        legend style={font=\scriptsize},
    ]
    
    \addplot [scatter, mark=o, nodes near coords, only marks, blue,
    point meta=explicit symbolic,
        every node near coord/.append style={anchor=west, xshift=2pt, font=\scriptsize},]
    table [meta=label] {
    x y label
    30900 0.945 AEGNN
    360000 0.931 EvS-S
    360000 0.915 NvS-S
    5100000 0.914 RG-CNNs
    4930000 0.902 G-CNNs
    840000 0.953 EV-VGCNN
    3690000 0.944 AsyNet
    2030000 0.927 YOLE
    };
    
    \addlegendentry{Software models}
    
    \addplot [scatter, mark=*, nodes near coords, only marks, blue,
    point meta=explicit symbolic,
        every node near coord/.append style={anchor=west, xshift=2pt, font=\scriptsize},]
    table [meta=label] {
    x y label
    4800 0.88 EvGNN
    };
    
    \addlegendentry{Hardware models}
    
    \addplot [scatter, mark=*, nodes near coords,only marks, red,
    point meta=explicit symbolic,
        every node near coord/.append style={anchor=west, yshift=2pt, xshift=2pt, font=\scriptsize},]
    table [meta=label] {
    x y label
    5200 0.907 EFGCN-S
    20200 0.923 EFGCN-L
    };
    
    \addplot [scatter, mark=*, nodes near coords,only marks, red,
    point meta=explicit symbolic,
        every node near coord/.append style={anchor=west, xshift=2pt, yshift=-4pt, font=\scriptsize},]
    table [meta=label] {
    x y label
    10600 0.920 EFGCN-B
    };
    
    \end{semilogxaxis}
    \end{tikzpicture}
\end{subfigure}

\caption{Classification accuracy versus number of parameters for the proposed EFGCN variants (red) on the N-Cars dataset, compared with prior event-based models implemented in software (blue circles) and in hardware (blue filled markers). The x-axis is logarithmic.}
\label{fig:model}
\end{figure}

{Our approach allowed us to significantly reduce the size of the model by up to 100 times, while limiting the accuracy loss to 2.2\% and 2.9\% for the N-Cars and N-Caltech101 datasets, compared to non-hardware state-of-the-art GCN methods.
The proposed custom EFGCN hardware module implemented for heterogenous ZCU104 SoC FPGA board achieves a~processing throughput of up to 13.3 million events per second with a~low latency of 4.39-9.31 ms.
The prepared solution supports multiple configurations (cf. Figure \ref{fig:model}).
Our approach offers strong scalability across various model sizes, input resolutions, and scene dynamics by leveraging highly parameterisable hardware modules. This design enables flexible adaptation of the system architecture to meet the specific requirements of a~given application.
These results present the potential of our approach for efficient real-time processing of DVS data in embedded systems.
Both the software model and the hardware modules have been published as an open source repository (\url{https://github.com/vision-agh/gcnn-dvs-fpga}).}

The remainder of this paper is organised as follows.
In Section \ref{sec:related} we present {existing work on event data processing with an emphasis on FPGA-implemented approaches.
Section \ref{sec:Method} details our proposed method for hardware-aware design of graph convolutional network architectures for event data processing. 
A~comprehensive description of the SoC FPGA implementation is provided in Section \ref{sec:FPGA}. 
In Section \ref{sec:Experiments} we present the experiments conducted and ablation studies.}
The paper concludes with Sections \ref{sec:limitations} and \ref{sec:summary}, where we present limitations, summarise our results and outline further work.

\begin{table*}[!t]
\centering
\caption{Comparison of graph network/convolution types and hardware platforms used for event data processing by graph neural networks.}
\resizebox{0.9\textwidth}{!}{%
\begin{tabular}{@{}lllll@{}}
\toprule
Work  & Year    & Network/Convolution type              & Task                      & Platform                  \\ \midrule
Sekikawa \cite{sekikawa2019eventnet} & 2019 & PointNet & Semantic segmentation and ego-motion & {CPU/GPU}                      \\
Wang \cite{wang2019space}  & 2019   & PointNet/PointNet++ & Gesture recognition       & {GPU}                       \\
Bi \cite{bi2019graph}     & 2019  & SplineConv                 & Classification               & Not mentioned             \\
Mitrokhin \cite{mitrokhin2020learning} & 2020 & GraphConv                      & 3D classification            & {GPU}   \\
Li \cite{li2021graph}    & 2021   & GraphConv                  & Classification               & {CPU}      \\
Schaefer \cite{schaefer2022aegnn} & 2022 & SplineConv                 & Classification and detection & {GPU}   \\
Gehrig \cite{gehrig2022pushing}  & 2022 & SplineConv              & Detection                 & {GPU}     \\
Jeziorek \cite{jeziorek2023memory} & 2023 & {PointNetConv}         & Classification and detection & {GPU}    \\

{Jeziorek \cite{jeziorek2024optimising}} & {2024} & {PointNetConv}         & {Graph generation and detection} & {GPU/FPGA}    \\

{Yang \cite{yan2020fpgan}} & {2024} & {PointNetConv} & {Classification} & {GPU/FPGA} \\

{This work} & {2025} & {PointNetConv} & {Classification} & {GPU/FPGA}

\\ \bottomrule
\end{tabular}%
}
\label{table:sota}
\end{table*}

\section{Related work}
\label{sec:related}

{Computer vision methods for traditional frame-based cameras are being developed for over 70 years.
However, with new vision sensors emerging, such as event cameras, these methods need to be altered to new types of data or completely new methods of data processing need to be established.
In this Section we focus on two different aspects: possible ways of processing data from an event camera and existing FPGA-based implementations.}

\subsection{{Event data processing}}

{Currently, three main approaches are employed for event data processing.
The first and most widely used approach involves \textbf{dense neural networks}, including convolutional neural networks \cite{6981783, messikommer2020event, perot2020learning, rebecq2019high, gehrig2019end, cannici2019asynchronous} and vision transformers (ViTs) \cite{Sabater_2022_CVPR, Gehrig_2023_CVPR, peng2023get}. These methods deliver high accuracy results and benefit from well-established techniques used for conventional image processing. However, they are not inherently designed for sparse and asynchronous events. To handle event data, these models first create dense representations like \textbf{event-frames} or \textbf{time-surfaces} \cite{afshar2020event}, which introduces additional latency. In this approach ‘empty' pixels are also processed, i.e. pixels without any events, which is inefficient in the case of sparse data. A~compromise is achieved through asynchronous methods \cite{gehrig2019end, cannici2019asynchronous}, which update only local pixels, where events occur. Nevertheless, these models tend to have large memory footprints and high computational complexity, while dense representations sacrifice high temporal resolution present in events.}

{The second approach utilises \textbf{spiking neural networks} (SNNs). They mimic the operation of brain neurons by accumulating membrane potential at each neuron and generating a~spike once a~certain threshold is exceeded. Since they are aligned with the event-based paradigm, SNNs can be directly applied to event cameras. However, the spiking mechanism inherent in SNNs is non-differentiable, making it difficult to apply the backpropagation approach for training \cite{lee2016training}. Additionally, the results often fall short in accuracy, and the learning mechanisms require further research to enhance their performance \cite{orchard2015hfirst, gehrig2020event, yao2021temporal, cordone2021learning, barchid2023spiking}.}

{The third and currently popular approach is to employ \textbf{graph convolutional networks} for the processing of event data. In this method, the data is represented as a~spatio-temporal graph, with each event corresponding to a~vertex connected to neighbouring vertices by edges. Table \ref{table:sota} compares various models and applications of GCNs for event data processing. Early work in this area demonstrated that GCNs significantly reduce the number of floating-point operations per event, due to their ability to maintain sparse data structures \cite{bi2019graph, mitrokhin2020learning, zhou2021event}. Moreover, later research demonstrated that they can be dynamically expanded by adding new events to the existing structure, with only local vertices requiring an update \cite{li2021graph, schaefer2022aegnn, gehrig2022pushing, gehrig2024low}.}

{A~notable contribution within this domain is the AEGNN framework \cite{schaefer2022aegnn}, which introduced the concept of k-hop graph updates, achieving high accuracy results with significantly reduced computational overhead compared to dense models. This approach has inspired further research \cite{jeziorek2023memory, jeziorek2024optimising, yanh_evgnn}, including our work. Moreover, since GCNs leverage standard backpropagation, this enables the training of larger models for tasks such as object detection \cite{schaefer2022aegnn, gehrig2022pushing, gehrig2024low}. Recent work \cite{gehrig2024low} presents a~hybrid system that integrates graph neural networks for event data with CNNs for frame data, highlighting the potential of GCNs for various computer vision applications. Overall, GCNs offer a~balanced solution, combining high accuracy output of dense models with the efficiency of SNNs.}

\changed{However, AEGNN's reliance on SplineConv (with learnable B-spline basis functions over continuous edge attributes) and k-hop bidirectional graph updates introduces high per-edge complexity, irregular memory access patterns, and unpredictable latency -- all challenging for an on-chip FPGA deployment. This motivated our use of PointNetConv with directed graphs, which avoids edge attributes and limits updates to the incoming vertex only (Sections~\ref{ssec:graph_gen} and~\ref{ssec:graph_conv}).}

\subsection{{FPGA implementations}}

{To fully exploit the advantages of event cameras, a~suitable computing hardware must be used.
FPGA platforms proved to be very efficient in traditional frame-based data processing with low latency \cite{garcia2014survey}.
Therefore, 
there has been a~growing interest in combining event-based data processing with FPGAs to develop energy-efficient vision systems.
The most comprehensive review in this area was conducted in 2024 \cite{kryjak2024event}, which surveyed various event-based FPGA solutions.}

{Classical approaches of event-based data processing on FPGA platform include HOTS \cite{lagorce2016hots} and PCA-RECT \cite{ramesh2020low} algorithms, which employ machine learning techniques like support vector machines (SVMs) combined with principal component analysis (PCA) for feature extraction. These methods require relatively small amount of hardware resources, but their application to complex computer vision tasks or larger datasets is limited. 

In case of AI-based methods, FPGA-based event processing can be divided into two subcategories: SNNs and CNNs.} {\textbf{SNN-based} solutions are characterised by their low latency \cite{camunas2018configurable, zhang2021fpga}, but the main drawback lies in the model sizes, as most SNN implementations use small input resolutions, typically around $32 \times 32$ or $64 \times 64$ pixels. As a~result, their quality rarely approaches state-of-the-art performance. On the other hand, \textbf{CNN-based} solutions \cite{linares2021dynamic, gao2024composable} can achieve high accuracy but require significant amount of hardware resources, exceeding the capacity of medium-sized FPGAs. This limits their applicability for edge devices and mobile robotics applications.}

{Besides the methods described, hardware implementations of \textbf{GCN-based} solutions have recently appeared.
Our previous work \cite{jeziorek2023memory} emphasised the limitations of existing graph convolution methods. This study revealed that most graph convolution layers require a~large number of parameters and often depend on edge attributes, which increase the graph's size. The most promising layer was found to be PointNetConv, which significantly reduces the number of parameters. It does not require edge attributes and improves the quality of the results.}

{Another limiting factor, which was not analysed in several GCN-based solutions, was graph generation procedure. An example is the work \cite{tao2022lw}, in which the authors focused on accelerating a~GCN network in hardware, using graph data generated earlier. Therefore, they did not face the challenges of handling raw input data in real time, which is usually crucial in practical situations.}

{In the study \cite{jeziorek2024optimising} we explored the hardware implementation of an efficient graph generation method for event data on the ZCU104 FPGA platform. This work introduced a~matrix-based approach for neighbour searches and demonstrated the advantages of using directed graphs, allowing asynchronous graph generation without the necessity of updating older vertices.}

{Recent paper \cite{yanh_evgnn} presents a~complete system for event-based graph processing on an FPGA. The authors adopted the same graph convolution method as in \cite{jeziorek2023memory} and utilised directed graphs, similar to the approach in \cite{jeziorek2024optimising, Dalgaty_2023_CVPR}. They also evaluated different neighbourhood search patterns. \changed{However, this solution differs from EFGCN in several important aspects. Without pooling, the graph size and edge count remain constant throughout the network, causing a~50$\times$ increase in FLOPs in deeper layers (Section~\ref{ssec:abl_pool}) and capping feature dimensionality at 32 elements per vertex. Moreover, the design relies on an off-chip DDR memory, introducing non-deterministic latency and higher energy cost (see Section \ref{ssec:hw_sota_compare}). In contrast, EFGCN introduces hardware 3D MaxPool layers that progressively reduce the graph between convolutions, enabling deeper architectures with up to 128-element feature maps, while operating entirely on-chip with deterministic latency.}

\changed{More broadly, existing FPGA-based GCN solutions either omit pooling layers entirely~\cite{yanh_evgnn}, rely on fixed pre-generated non-event graphs~\cite{wang2022bignn, tao2022lw, zhou2022inference}, or depend on an external off-chip memory~\cite{yanh_evgnn}. To the best of our knowledge, no prior hardware implementation supports all characteristic GCN layers~-- including pooling~-- within a~single on-chip pipeline, nor has 3D MaxPool for event data been previously realised in hardware. EFGCN directly addresses both gaps.}


\section{The proposed method}
\label{sec:Method}

Our research was motivated by the need to bridge the gap between the hardware implementation of GCNs and their application in event data processing. 
For this purpose, we emphasise the sparse nature of the input data, particularly its temporal aspect, and incorporate it directly into the network design.
The goal was to develop a~solution that maximises the use of information captured by event cameras, enabling accurate and instant event-by-event data processing (as soon as the data is registered by a~neuromorphic sensor) with minimal latency, low computational load and low resource utilisation. 
In this way, we aim to design a~system suitable for real-time mobile robotics applications operating in highly dynamic environments.
In this Section, we outline the solutions and necessary modifications used for implementing a~GCN on a~SoC FPGA.

\subsection{Event-graph construction}
\label{ssec:graph_gen}

Event cameras are characterised by the ability to capture changes in brightness at the individual pixel level, which distinguishes them from traditional frame-based cameras. The operation of an event generation is governed by a~threshold mechanism, which determines whether the change in light intensity for a~specific pixel exceeds a~predefined value. The outcome of this process is an event stream that can be described as a~sequence of tuples \(ev_i = (u_i, t_i, p_i)\), where \(u_i = (x_i, y_i)\) indicates the pixel location at which the event was generated, \(t_i\) is the timestamp, and \(p_i \in \{-1, 1\}\) denotes whether the change was negative or positive, respectively.

In the literature \cite{bi2019graph, schaefer2022aegnn}, a~\textbf{standard method} for generating a~graph from events is to construct a~spatio-temporal graph \( \mathcal{G} = (\mathcal{V}, \mathcal{E}) \), where each event \(ev_i\) is represented as a~vertex \( v_i \in \mathcal{V}\) with a~position \( \mathcal{P}_i = (u_i, t_i) \) and attribute \(\mathcal{X}_i = (p_i)\). Subsequently, for each pair of vertices \( i \) and \( j \), an edge \( e_{ij} \in \mathcal{E}\) is generated, if their Euclidean distance \(d_{ij}\) represented as:

\begin{equation}
d_{i,j} = \sqrt{(x_i - x_j)^2 + (y_i - y_j)^2 + (t_i - t_j)^2}
\label{eq:distance}
\end{equation}

\noindent is smaller than a~predefined parameter \( R \). To normalise the time dimension for better comparability with spatial coordinates, the timestamps \( t_i \) are scaled by a~factor \( \alpha \), usually equal to the time window, resulting in \(t_i^* = t_i \cdot \alpha\). Additionally, the number of edges is limited to \( D_{max} \) per vertex to avoid excessive edge generation. \changed{In practice, \( D_{max} \) defines a~trade-off between capturing more complete local spatio-temporal relations and maintaining bounded memory usage and computational cost, which is particularly important for efficient hardware implementation.}

{However, directly searching for neighbouring vertices across the entire graph presents computational and memory challenges, especially in the context of hardware implementations, which would require using external memory, resulting in increased latency and power consumption. Traversing all vertices in a~graph of millions of events results in \( O(n) \) complexity. Our objective is to minimise the number of searches per event to maintain real-time performance. Moreover, generating directed edges that support the continuous addition of new events without requiring updates to the existing graph is crucial for efficient hardware processing, as discussed further in Subsection \ref{ssec:graph_conv}.}

{To address these challenges, we adopt the method proposed in \cite{jeziorek2024optimising}, which optimises graph generation for hardware. For a~comprehensive explanation of the method, refer to the original work. Below, we summarise the key steps of the approach.}

{Building on prior works \cite{li2021graph, schaefer2022aegnn} and classical CNN-based models, the method begins by normalising both the spatial \( u_i = (x_i, y_i)\) and temporal \( t_i \) coordinates to a~common range \( \beta \). The normalised coordinates are then discretised to integer values:}
\begin{equation}
x_i^* = \left\lfloor \beta \cdot \frac{x_i}{\changed{X}} \right\rfloor, \quad y_i^* = \left\lfloor \beta \cdot \frac{y_i}{\changed{Y}} \right\rfloor, \quad t_i^* = \left\lfloor \beta \cdot \frac{t_i}{T} \right\rfloor.
\label{eq:normalization}
\end{equation}

\noindent \changed{Here, \(X\) and \(Y\) denote the sensor's spatial resolution (width and height)} and \(T\) is the time window. This process maps the vertices into the set \( \mathcal{V} \subset \mathbb{N}^3 \), where each component ranges from \( 0 \) to \( \beta \).

{Neighbour searches are performed using a~\textbf{neighbourhood matrix} (shortly $NM$), a~two-dimensional matrix that corresponds to the normalised spatial resolution of the events. Each cell in the $NM$ stores the timestamp and the polarity of the most recent event at given spatial coordinates. For each new event \( ev_i \), potential neighbours are searched among the pixels within a~radius \( R \) of its position \( u_i^* \). If the timestamp \( t_j^* \) of a~neighbouring pixel \( u_j^* \) satisfies Eq. \eqref{eq:distance}, a~direct edge \( e_{ij} \in \mathcal{E}\) is created and the timestamp in the matrix is consequently updated to \( t_j^*\).}

{This approach allows for the initialisation of a~fixed-size data structure (\( \beta \times \beta \)), that stores information about previous events, limits the number of neighbour searches based on the parameter \( R \), and enables the continuous processing of events while maintaining directed edges. \changed{In our implementation, \( D_{max} \) is not defined as a separate hyperparameter; instead, it is implicitly bounded by the neighbourhood search radius \( R \) (set to 3 in our experiments) and the spatial discretisation, which together limit the maximum number of edges per vertex to 29.} The applied \(\beta\) parameter with spatial resolution \changed{\((X, Y)\)} and time window \(T\) for different datasets used for the evaluation of our method are shown in Section \ref{sec:Experiments}.} 

\subsection{Graph convolution}
\label{ssec:graph_conv}

{The core operation in \textbf{graph convolutional networks} is a~convolution, which facilitates the efficient propagation of information between locally connected vertices. This enables updating vertex representations and extracting relevant features. Graph convolution generally involves three stages:} 

1. \textbf{Message Function (\(\phi\))}:  
   in the first stage, the message function \(\phi\) operates on a~vertex \(v_i\) and its neighbours \(v_j \in \mathcal{N}(i)\), determining the information exchange between these vertices. The message function can utilise the vertex attributes \(\mathcal{X}_i\), \(\mathcal{X}_j\), their positions \(\mathcal{P}_i\), \(\mathcal{P}_j\), and edges \(\mathcal{E}\).

2. \textbf{Aggregation Function (\(\bigoplus\))}:  
   next, the aggregation function \(\bigoplus\) combines the message values collected from all neighbouring vertices, producing a~representative value that includes the information from the local neighbourhood of vertex \(v_i\).

3. \textbf{Update Function (\(\gamma\))}:  
   finally, the update function \(\gamma\) adjusts the vertex attribute \(\mathcal{X}_i\) using the aggregated information. 
   
The overall message passing process can be mathematically represented as Eq. \eqref{eq:general_conv}.

\begin{equation}
   \hat{\mathcal{X}_i} = \gamma \left(\bigoplus_{j \in \mathcal{N}(i)} \phi(\mathcal{X}_i, \mathcal{X}_j, \mathcal{P}_i, \mathcal{P}_j, \mathcal{E}) \right)
   \label{eq:general_conv}
\end{equation}

\noindent Here, both \(\phi\) and \(\gamma\) are differentiable functions, typically implemented as multi-layer perceptrons (MLPs). The aggregation function \(\bigoplus\) can be a~simple operator like sum or maximum.

{As shown in Table \ref{table:sota}, several different convolutional layers are available and widely used, such as SplineConv \cite{schaefer2022aegnn, gehrig2022pushing, bi2019graph}, GraphConv \cite{mitrokhin2020learning, li2021graph}, or network architectures like PointNet and PointNet++ \cite{sekikawa2019eventnet, wang2019space}. However, these solutions often have high memory demands \cite{jeziorek2023memory}, which is not a~major issue when using GPUs, but becomes critical in embedded implementations. The extensive memory utilisation by the convolutions has a~potential to adversely impact the reduction in dimensionality or depth of the network, which leads to a~loss in quality. Therefore, in our work, we focused on using the \textbf{PointNetConv}\footnote{The implementation of the PointNetConv layer presented here is based on its implementation in the PyTorch Geometric library \cite{Fey2019PyTorchGeometric}, which is based on the work \cite{qi2017pointnet++}.} layer.}

This choice was guided by the results of our prior research \cite{jeziorek2023memory}, where we achieved a~major reduction in model size and its inherent suitability for a~point cloud data.
The {PointNetConv} layer is designed to efficiently process vertex features in 3D point clouds, implementing the transformation defined by Eq. \eqref{eq:pointnetconv}:

\begin{equation}
\label{eq:pointnetconv}
\hat{\mathcal{X}_i} = \gamma \left (\underset{j \in N(i)}{\max} \phi (\mathcal{X}_j, \mathcal{P}_j - \mathcal{P}_i)\right ),
\end{equation}

\noindent where \(\phi\) is a~function that processes the neighbour vertex attribute \(\mathcal{X}_j\) with the relative spatial coordinate \(\mathcal{P}_j - \mathcal{P}_i\), and the \(\max\) operator selects a~representative attribute based on information received from the neighbours \(\mathcal{N}(i)\). In our implementation, the message function \(\phi\) is a~simple linear transformation, mapping input features from dimension \(in+3\) to \(out\), where the additional \(+3\) accounts for the spatio-temporal dimensions of the events. To minimise complexity, we omit the update function \(\gamma\), and in order to apply the BatchNorm merging \cite{jacob2018quantization}, we used normalisation after \(\phi\) function.

\changed{The selection of PointNetConv as the convolution method for the hardware-implemented system was therefore motivated by its limited computational complexity, wherein the features of vertices connected by an edge are combined using a simple max() function. The efficiency of PointNetConv enables a~reduction in model size, which in turn decreases the demand for internal FPGA memory, which is highly constrained and energy-intensive resource.}
{We investigated the effect of the convolution used on the model size and the results in Section \ref{ssec:ablation_conv}.}

\subsubsection*{\textbf{Edge directionality}}

{In the case of graphs that are \textit{bidirectional} or \textit{directed against time}, each new event added to the graph can be connected to the neighbourhood \(\mathcal{N}(i)\) of an existing vertex. This necessitates recalculating not only the attribute of the new event, but also the attributes of all vertices it is connected to. Such an update propagates further into the graph since each subsequent layer utilises information from more distant neighbours. This process, described in detail in \cite{schaefer2022aegnn} as a~\textbf{k-hop neighbourhood}, presents challenges for hardware implementations. Specifically, it requires storing information about previous vertices for each convolution and updating them continuously.
In contrast, for graphs \textit{directed with time} new events are connected only to the nearest vertices, and only the new vertex needs an update. This significantly reduces the number of operations per event \changed{while enabling asynchronous graph updates. This mechanism is crucial for efficient hardware implementation, as it can exploit the spatio-temporal sparsity of event-based data.}}

\subsection{Graph MaxPool}
\label{ssec:maxpool}


Graph MaxPool is a~technique used to reduce the number of vertices in a~graph. It is particularly useful in deeper layers of neural networks, where the size of attributes can increase significantly. The technique involves partitioning the data space into uniform \(\mathcal{C}_k\) clusters of size \(g_x \times g_y \times g_t\). For each of these non-empty clusters, a~new vertex is selected whose attribute value corresponds to the maximum attribute value among all vertices in the cluster, as represented by Eq. \eqref{eg:maxpool}.

\begin{equation}
\label{eg:maxpool}
    \mathcal{X}_k = \max_{i \in \mathcal{C}_k} \mathcal{X}_i
\end{equation}

Meanwhile, the position of the new vertex is determined as the average of the positions of all vertices in the cluster, as in Eq. \eqref{eq:claster_pos}.
\begin{equation}
\label{eq:claster_pos}
\mathcal{P}_k = \frac{1}{|\mathcal{C}_k|} \sum_{i \in \mathcal{C}_k} \mathcal{P}_i
\end{equation}

In this process, connections that link vertices from different clusters are merged into a~single edge between new points, eliminating redundant connections and those internal to the groupings.

{In the works \cite{schaefer2022aegnn, gehrig2022pushing}, the MaxPool operation is performed two-dimensionally with respect to space, meaning each cluster has a~\(g_t\) parameter set to the maximum value. As a~result, for a~given spatial range \(g_x \times g_y\), there can be only one output vertex. This was motivated by computational reduction due to a~significant decrease in the number of vertices. However, there are two key issues with this approach. The first is the continuous change in the spatial structure of the graph after pooling, as with each new event \(ev_i\), the position \(\mathcal{P}_k\) is modified according to Eq. \eqref{eq:claster_pos}. The second issue, highlighted by the authors in \cite{gehrig2022pushing}, is that this operation can cause bidirectional edges between output nodes. Both effects impact the continuous updating of the graph itself as well as the recalculation of vertex attributes, which in case of an FPGA results in a~constant need to read, recompute, and write data.}

{To address this, we propose two modifications to this method to enable its hardware implementation. The first modification is to apply 3D pooling, where the cluster size is set to the same value for each dimension, so \(g_x = g_y = g_t = g\). The result is a~larger number of vertices compared to the 2D method, but the pooling operation is divided into time intervals, which negates the need to update vertices in the temporally older parts of the graph. The second modification is to scale down the vertex positions according to the cluster size, so that each position value \(\mathcal{P}_k\) is divided by the parameter \(g\). This ensures that the edges in the output graph remain directed with time and reduces resource consumption on an FPGA.}

{Our analysis in Section \ref{ssec:abl_pool} shows that using 3D pooling positively affects the qualitative performance.
Also, scaling the graph down causes a~significant reduction in both memory and logic resources (see Sections \ref{ssec:MaxpoolHW} and \ref{ssec:sync_conv}).}

\section{Hardware implementation}
\label{sec:FPGA}

After extensive ablation studies in software (Section \ref{ssec:ablation}), we developed a~set of hardware modules for processing an event stream using graph convolutional neural networks.
{Our custom architecture implemented in SystemVerilog was simulated for consistency with the software model and evaluated across multiple configurations on the Zynq UltraScale+ MPSoC ZCU104 Evaluation Kit (featuring the XCZU7EV-2FFVC1156 chip).}

The hardware architecture was designed as a~pipeline of modules implementing successive layers of a~graph convolutional network that process data in a~completely parallel manner -- each layer can process data simultaneously, as long as its input data is available (Fig. \ref{fig:hw_schema}). 
\changed{The hardware-aware optimizations proposed in Section \ref{sec:Method} were aimed, among other objectives, at reducing memory requirements, thereby enabling the exclusive use of internal memory resources during inference (no off-chip memory). Satisfying this constraint not only reduces energy consumption (as the use of BRAM/URAM is more energy-efficient), but also ensures constant and deterministic memory access latencies. Consequently, the computational architecture could be designed to fully exploit the spatio-temporal sparsity of event-based data by appropriately tailoring the degree of computational parallelism.}
The system consists of two parts -- asynchronous (operating in an event-by-event manner) and synchronous, where operations are performed on sub-graphs.
The toolchain can process sequence of events of any length continuously, generating a~prediction at the output for data recorded during the last \texttt{TIME\_WINDOW} (duration of a~single sample of event data sufficient to perform the classification -- in described configurations we support windows of 50 and 100 ms), but 4 times as \changed{often} (features available at system's output every \texttt{TIME\_WINDOW}/4).

The system assumes the use of a~heterogenous SoC FPGA platform, where feature extraction is implemented in the programmable logic (PL) and the network head is performed in the processing system (PS).
{The operating frequency of 200 MHz was selected based on experimental evaluation to balance latency and timing requirements.}
Lower clock frequencies could be used to reduce power consumption at the cost of increased latency.
In the following sections, we present the functional description of each hardware module, preserving their order in the system pipeline. 

\begin{figure}[!t]
    \centering
    \includegraphics[width=0.8\columnwidth]{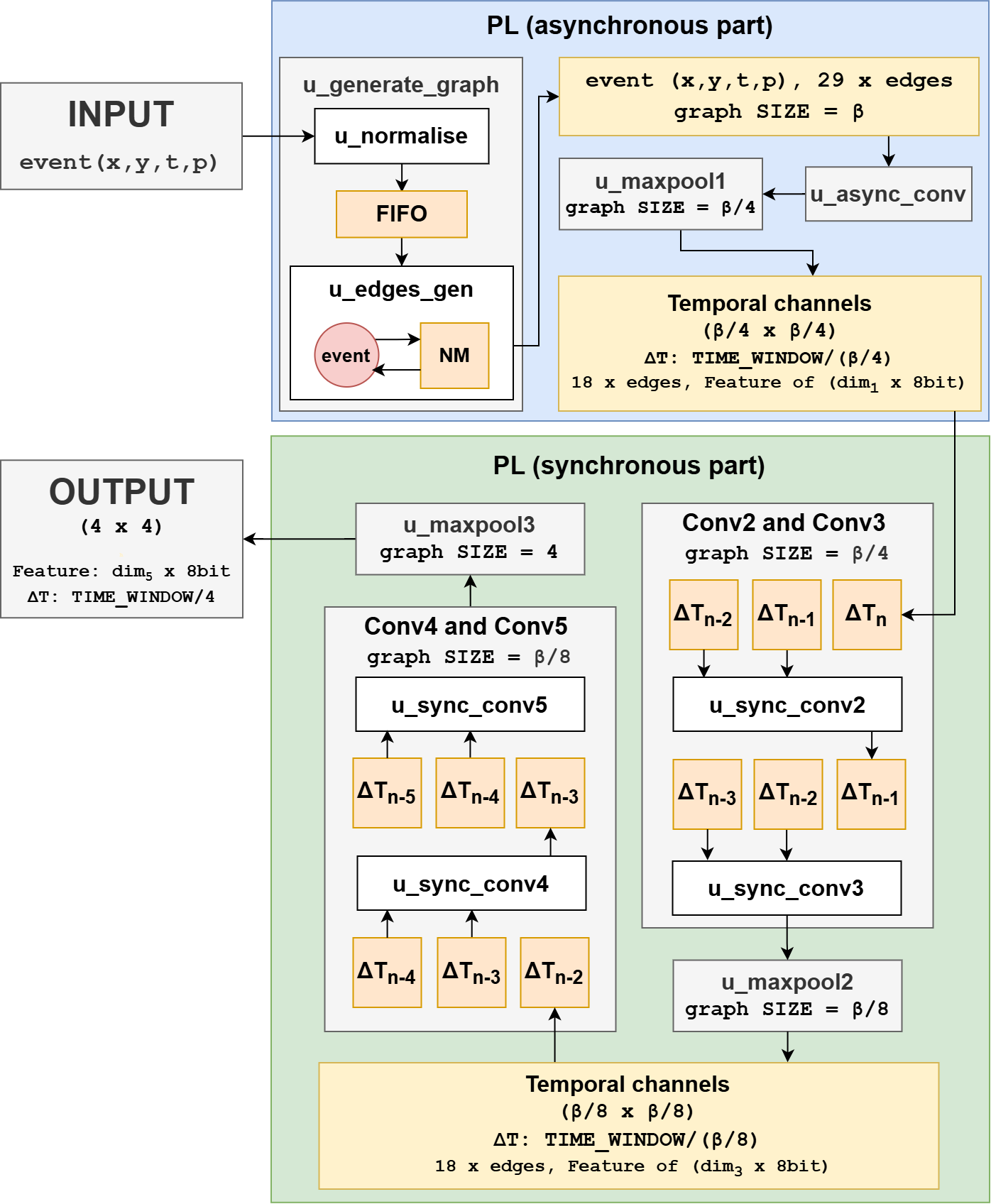}
    \caption{{Schematic of the EFGCN network hardware modules with the asynchronous part marked in blue and the synchronous in green. The orange modules represent integrated memory, while the characteristics of data transferred between the selected modules are highlighted in yellow blocks. The number of features for a~particular graph convolution layer $dim_{1}$-$dim_{5}$ depends on the model (Small, Base or Large). We support \changed{configurations with normalisation size} \( \beta \) $= 128$ and \( \beta \) $= 256$.}  }
    \label{fig:hw_schema}
\end{figure}


\subsection{Graph generation (\texttt{u\_generate\_graph})}
{For testing purposes, in the designed system consecutive events are read by the processor from an SD card and then transmitted to the programmable logic in such a~way as to maintain the intervals between the events, corresponding to their timestamps (emulating the actual output of a~DVS). We also successfully evaluated the continuous processing of data read directly from a~DVS integrated with the PL \changed{\cite{wzorek2025live}}}.

{The events are then normalised in the FPGA (Eq. \eqref{eq:normalization}) -- the coordinate values $x$, $y$ and the timestamp $t$ are scaled to \( 0 \)-\( \beta \) (graph size) range. 
Currently, we support graphs sizes of $128 \times 128 \times 128$ and $256 \times 256 \times 256$.
The polarisation value is encoded as a~single bit.
The normalised events are written into a~FIFO queue implemented with BlockRAM to be then read by the following modules.
This buffering mechanism ensures correct processing when the dynamics of the observed scene exceed the system’s bandwidth (13.3 million events per second that can be processed by our architecture).}

{The subsequent events read from the FIFO are used to update the graph representation.
For this purpose, it is necessary to determine the list of neighbours for the currently processed vertex.

\changed{The hardware-aware optimisation techniques described in Section \ref{ssec:graph_gen} enable the implementation of a~graph generator with reduced memory resource utilisation. The use of a~directed graph allows candidate edge connections to be searched exclusively among previously processed vertices. Furthermore, limiting the maximum number of connected vertices sharing the same 
$x$ and $y$ coordinates ensures that edge construction requires storing only information related to the most recent events within each memory cell.}
The \textbf{neighbourhood matrix} (Section \ref{ssec:graph_gen}) is implemented as a~two-port BlockRAM memory of \( \beta \) $\times$ \( \beta \) depth and width of \( \beta \)$+2$ (storing the timestamp of the last recorded event for a~particular pixel, its polarity and \texttt{is\_empty} flag).
For each event read from the FIFO, the edge candidates are read from the $NM$ -- 29 surrounding values (for $R=3$).
For each read, the semi-sphere condition is checked (Eq. \eqref{eq:distance}).
Next, the currently processed event is written to the $NM$.
This operation, due to the use of dual-port memories, requires 15 READ operations on one of the ports, and 14 READ operations and one WRITE operation on the other.

The latency of 15 clock cycles for a~single event is a~computational bottleneck of the proposed system. 
It can be determined that, with a~200 MHz clock, the throughput of the proposed system is 13.3 MEPS,} which exceeds the values observed for real event stream data in the utilised datasets (1.35 MEPS for N-Caltech101, 0.59 MEPS for N-Cars, 0.34 MEPS for CIFAR10-DVS, and 0.063 MEPS for MNIST-DVS).

Once the $NM$ analysis is complete, the module outputs pairs of events ($x$, $y$, $t$, and polarity), and a~vector describing their edges (including polarity of connected events).

\subsection{Asynchronous convolution (\texttt{u\_async\_conv})}

Successive pairs of events and their edges appear at the first convolution module's (Section \ref{ssec:graph_conv}) input asynchronously with preserved order -- no more frequently than once every 15 clock cycles, but with an unknown time span (depending on the dynamics of the observed scene).

For each processed vertex, the graph convolution {(implementation of Eq. \eqref{eq:pointnetconv}) requires performing maximum of 30 multiplications -- one for its own feature map (self-loop) and one for the feature map of each neighbouring vertex connected by an edge (a~maximum of 29).
The input feature map consists of four components: polarisation and the positional difference between connected vertices (for the self-loop, this is $(0,0,0)$).
No extra memory is required, since the described initial convolution process only the values that are already present at the module’s input.}

{The implementation of the linear function $\phi$ (Eq. \eqref{eq:pointnetconv}) utilises two parallel modules for matrix multiplication, which are realised in LUT logic resources.
For the first convolutional layer, all elements of the output vector are computed in parallel -- therefore, the processing time for a~vertex-edge pair equals 15 clock cycles.
For each resulting value (after adding the bias), a~requantisation process is required.
It consists of scaling (using DSP and bit-shifting) followed by the addition of a~constant \texttt{ZERO\_POINT}.}
The resulting vectors for each edge (and self-loop) are compared with an element-wise maximum operation (taking into account layer-specific minimum values -- ReLU activation).
This generates the final feature map vector, which is propagated to subsequent model layers. The output of the module is the currently processed event, a~list of its edges and a~vector of its features.

\changed{It should be noted that for this convolution (unlike the subsequent ones, see Section \ref{ssec:sync_conv}), the entire matrix-vector multiplication is performed in parallel in order to avoid introducing an additional bottleneck -- the module must provide throughput not lower than that of the graph generation module. Consequently, the number of output features of the first convolution in each of the evaluated models (see Section \ref{subsec:model}) is limited -- increasing it would result in higher utilisation of both DSP and logic resources.}

\subsection{3D MaxPool (\texttt{u\_maxpool})}
\label{ssec:MaxpoolHW}

\begin{figure}
    \centering
    \includegraphics[width=0.6\columnwidth]{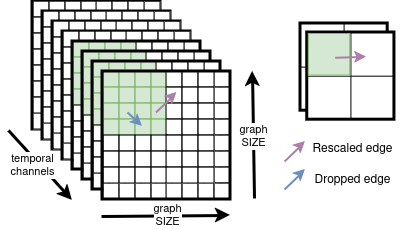}
    \caption{Operation of the 3D MaxPool layer. The green area  refers to the pooling region before and after the $\max$ \changed{operation}.}
    \label{fig:MaxPool}
\end{figure}

{Immediately following the first convolution, a~3D MaxPool operation (Section \ref{ssec:maxpool}) with a~$4 \times 4 \times 4$ kernel is applied.
Its purpose is to scale the graph in each of the three dimensions -- x, y, and time -- by replacing the input events within $4 \times 4 \times 4$ regions with a~single output vertex.
Consequently, the graph of \texttt{SIZE} (for the input graph \texttt{SIZE =} \( \beta \)) is replaced for subsequent layers with a~graph of \texttt{SIZE/4}.
The output feature vectors are the results of an element-wise maximum operation (Eq. \eqref{eg:maxpool}; visualised on Fig. \ref{fig:MaxPool}) applied to all inputs within a~given region.
This operation leads to the following:
\begin{itemize}
\item The number of processed vertices for subsequent layers in the network model is reduced 16 times, leading to a~significant decrease in the required memory resources.

\item The number of edges is also reduced. After $4 \times 4$ clustering, the maximum positional difference between connected vertices is reduced to $R=1$, and their number is reduced to 17. As a~consequence, the maximum number of \changed{matrix-vector} multiplications required for graph convolution in subsequent layers is reduced as well (from 30 to 18). \changed{As multiplication accounts for a~significant portion of the utilised logic resources, this effect is crucial for the scalability of the hardware architecture.}

\item The processing order of events is disrupted.
Scaling the graph not only among the x and y coordinates, but also in the time plane, forces data accumulation within specific intervals.
Subsequent layers can only process data once the accumulation is complete.

\end{itemize}

The hardware module of a~{3D MaxPool} operates on the BlockRAM \textit{feature memory} with a~depth of \texttt{(SIZE/4)}$^2$, which is addressed using x and y coordinates.
Subsequent events are assigned to one of its cells, where the feature map is stored along with an updated list of its edges.
A~single read and write operation of a~given cell allows the execution of the element-wise $\max()$, when more than one event is assigned to it.
Event accumulation is performed within a~time unit:
\begin{equation}
\Delta T = \frac{TIME\_WINDOW}{(SIZE/4)}.
\label{eq:maxpoolhw}
\end{equation}

As a~result, the \textit{feature memory} contains the complete feature map after a~$\Delta T$ period defined by Eq. \eqref{eq:maxpoolhw}, and only then can it be processed by subsequent layers.
The disruption of event order forces a~change in the data processing method in the next part of the model (referred to as ‘synchronous') -- instead of event-by-event processing, two-dimensional feature maps called \textit{temporal channels} are processed, representing all events that occurred within a~given $\Delta T$.

An advantage of using the 3D MaxPool module \changed{(along with constant-latency memory resources)} is the ability to precisely determine the required throughput for each subsequent layer in the synchronous part of the system.
Based on this value, an appropriate computation strategy is chosen to ensure that as many operations as possible can be executed sequentially.
}

\subsection{Synchronous graph convolution (\texttt{u\_sync\_conv})}
\label{ssec:sync_conv}

\begin{figure}
    \centering
    \includegraphics[width=0.85\columnwidth]{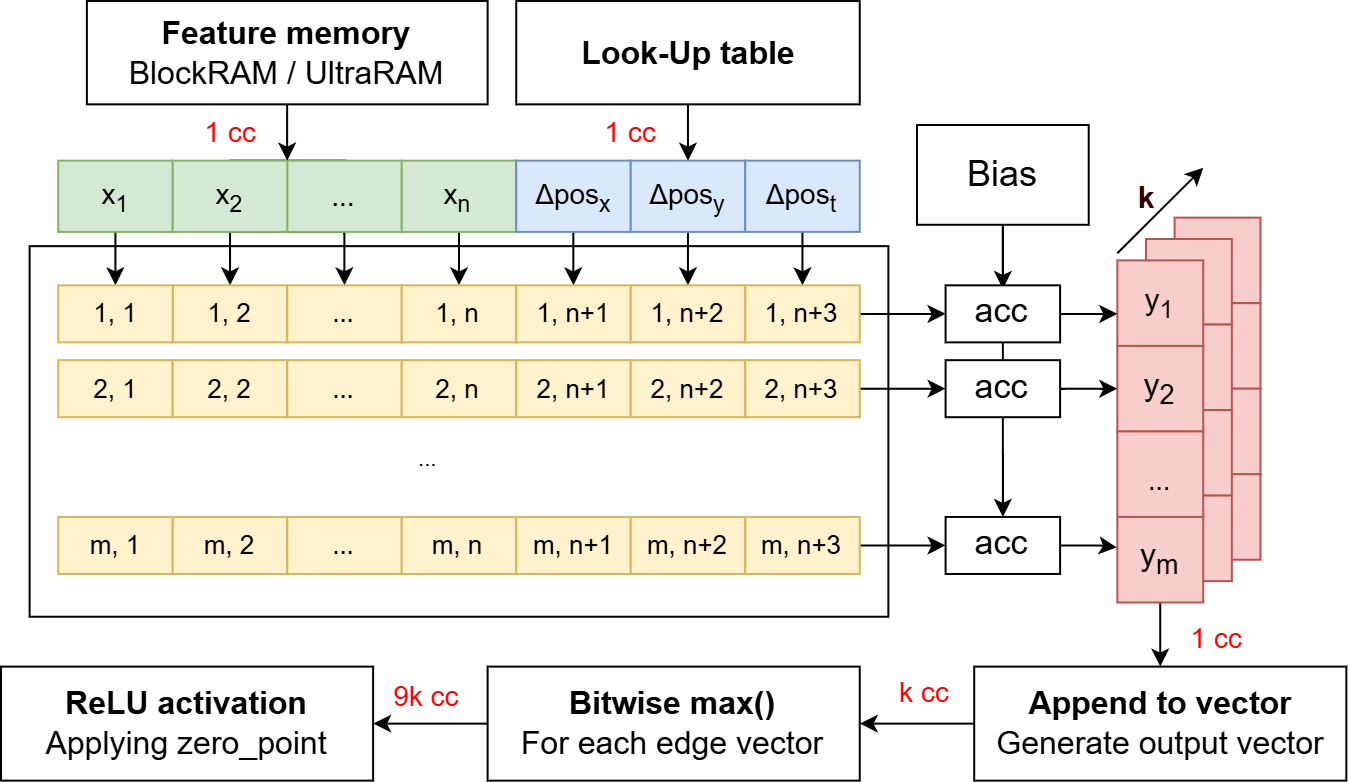}
    \caption{{The graph convolution module in the synchronous part of the system. The input vector, composed of features retrieved from memory (green) and position differences (blue), is processed by the multiplier matrix (yellow). The resulting vector (red) is generated through an append operation in $k$ steps, with $m$ elements each.}}
    \label{fig:sync_conv_mul}
\end{figure}

{To process the entire \textit{temporal channel}, it is necessary to compute the output of the linear function $\phi$ (Eq. \eqref{eq:pointnetconv}).
For each element, the system must read the feature maps of the currently processed vertex as well as all vertices connected to it by an edge, resulting in a~maximum of 18 memory accesses.
Notably, eight of these potential edges connect vertices within the same $\Delta T$, while the remaining nine connect to neighbouring vertices from the previous $\Delta T$ ($R=1$ along the temporal axis as well).
Executing this convolution requires access to the last two \textit{temporal channels}, a~constraint that was taken into account when designing the ‘feature memory' mechanism (Fig. \ref{fig:FeatureMem}).

The computation of the linear function $\phi$ for a~single vector $x_{i}$ consists of the following (Fig. \ref{fig:sync_conv_mul}):

\begin{itemize}
\item The vector $x_{i}$ is constructed by aggregating an n-element feature vector (retrieved from ‘feature memory') and three position difference values (Eq. \eqref{eq:pointnetconv} -- \(\mathcal{P}\)). The position differences are obtained from a~look-up table (LUT) based on the coordinate differences between vertices connected by an edge.

\item Each element of the output vector $\phi(x_{i})$ is computed by multiplying  $x_{i}$ by a~weight vector. Depending on the required module throughput (determined by $\Delta T$), the number of parallel vector multipliers ($m$) can be adjusted.
The more multipliers are used, the higher the resource consumption. Therefore, the smallest possible value of $m$ to ensure that the required throughput is maintained without exceeding resource constraints is selected.

\item For each iteration (i.e. in each clock cycle), $m$ elements of the output feature vector are determined. This process is repeated $k$ times for a~single input vector to obtain the $\phi(x_{i})$ of dimensions $dim = m \times k$.

\end{itemize}

With the required access to two preceding \textit{temporal channels}, which are stored in independent ‘feature memory' units, the system processes two vectors simultaneously by default.
The convolution result for a~given vertex is determined as the outcome of the bitwise $\max$ operation applied to all $\phi(x_{i})$ vectors.
The number of clock cycles ($CC$) required for this computation can be expressed as:
\begin{equation}
CC_{vertex} = 9 \times \frac{dim}{m},
\label{eq:cc_vertex}
\end{equation}
where $dim$ represents the size of the output feature map, and $m$ denotes the number of parallel multipliers used.
In summary, the processing of an entire \textit{temporal channel} takes:
\begin{equation}
CC_{t} = CC_{vertex} \times SIZE  \times SIZE
\label{eq:cc_t}
\end{equation}
where $SIZE$ represents the current size of the processed graph, which decreases after each successive MaxPool. 

The strategy for reducing LUT resource utilisation involves selecting the maximum number of parallel multipliers $m$ so that $\Delta T_{cc}$  (in clock cycles) satisfies Eq. \eqref{eq:cc_condition}.
\begin{equation}
\Delta T_{cc} \geq 9 \times \frac{dim}{m} \times SIZE  \times SIZE
\label{eq:cc_condition}
\end{equation}

In the synchronous part of the system, each graph convolutional layer can employ a~different multiplication strategy (for decreasing $SIZE$ with each MaxPool).}

\subsection{Memory management}

{For GCN inference, it is necessary to store intermediate feature maps (in the synchronous part) as well as weights.
In the proposed hardware module, only internal memory resources that offer constant and limited latency (BlockRAM and UltraRAM) are utilised -- access to \changed{any off-chip} memory is not required.
To minimise memory resource usage, a~memory-sharing mechanism has been implemented between successive layers (Fig. \ref{fig:FeatureMem}).}

\begin{figure}
    \centering
    \includegraphics[width=0.6\columnwidth]{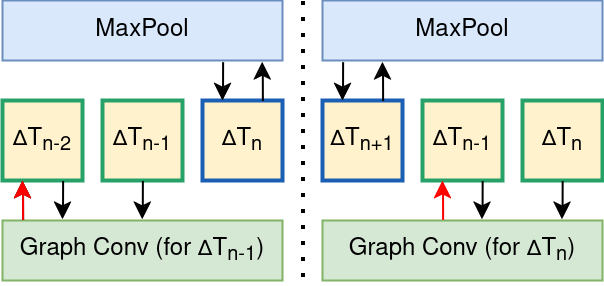}
    \caption{The memory switching method -- memory blocks are represented as yellow squares, while arrows indicate memory writes and reads. The red colour denotes memory reset.}
    \label{fig:FeatureMem}
\end{figure}

{Each module implementing the~\textit{feature memory} consists of three independent memory blocks (implemented with BlockRAM or UltraRAM, depending on the feature vector size).
While the preceding layer (e.g. MaxPool) writes data to one of these memory blocks generating a~\textit{temporal channel} for $\Delta T_{n}$, the other two (storing data for $\Delta T_{n-1}$ and $\Delta T_{n-2}$, respectively) are used by the next layer.
Once convolution processing for an entire \textit{temporal channel} is completed, the oldest memory block (corresponding to $\Delta T_{n-2}$) is reset.
An appropriate multiplication strategy ensures that the graph convolutional layer completes processing before the next \textit{temporal channel} is ready. At that point, memory switching occurs so that new data is stored in the recently cleared memory block.

Weight matrices are stored in distributed RAM for the first (asynchronous) convolutional layer due to the requirement of a~simultaneous access and the relatively small size. In the synchronous part of the network, the weight matrices are stored in BlockRAM. However, the memory utilised for this purpose represents only a~small portion of the overall usage.}

\subsection{Hardware module's output}

{Regardless of the chosen configuration of the GCN architecture, the processed graph at the output of the reconfigurable part of the system (after the sequence of MaxPool layers and graph convolution) has a~size of \texttt{SIZE} $= 4$.

As a~result, successive \textit{temporal channels} are transmitted to the software part of the heterogeneous system every \texttt{TIME\_WINDOW}$/4$ ms. 

The final feature map is used as input to the linear layer implemented in the PS to determine the classification results.
We implemented the head of the network in the PS to increase the versatility of the proposed architecture -- the same hardware module for feature extraction can be used for different computer vision tasks (while it could be implemented in reconfigurable logic, with relaxed throughput requirements, this is not necessary).

\changed{The effective inference rate remains independent of the network architecture or event rates and depends exclusively on the \texttt{TIME\_WINDOW}}:
considering continuous event data processing, predictions can be generated every \texttt{TIME\_WINDOW}$/4$ ms, with each prediction taking into account the most recent \texttt{TIME\_WINDOW} ms of data.
\changed{The prediction latency of the system is influenced by the chosen architecture (the number of features in each convolution layer and the adopted parallel multiplication strategy); its evaluation is presented in Section \ref{ssec:HW-eval}.}}

\section{Experiments}
\label{sec:Experiments}
\subsection{Setup}

\subsubsection{Datasets} In our experiments, we focused on the object classification from events. We selected four commonly used datasets including N-Cars \cite{sironi2018hats}, N-Caltech101 \cite{orchard2015converting}, CIFAR10-DVS \cite{CIFARDVS}, and MNIST-DVS \cite{6407468}. The details about the dataset statistics and the parameters used during graph generation are summarised in Table \ref{table:datasets}. For the N-Caltech101 dataset, due to the larger resolution and dynamics, the normalisation parameter \( \beta \) is set to 256 and \texttt{TIME\_WINDOW} to 50 ms.

\begin{table}[!t]
\centering
\caption{Details of statistics and parameters in the graph generation stage for the event datasets used, including N-Cars (N-C), N-Caltech101 (N-Cal), CIFAR10-DVS (C-DVS) and MNIST-DVS (M-DVS).}
\resizebox{0.7\columnwidth}{!}{%
\begin{tabular}{@{}ccccc@{}}
\toprule
Datasets            & N-C \cite{sironi2018hats}   & N-Cal \cite{orchard2015converting} & C-DVS \cite{CIFARDVS} & M-DVS \cite{6407468} \\ \midrule
Samples             & 24029     & 8246         & 10000       & 30000     \\
Classes             & 2         & 100          & 10          & 10        \\
Duration            & 100 ms    & 300 ms       & 1280 ms     & 2-3 s   \\
Resolution & $120 \times 100$ & $240 \times 180$      & $128 \times 128$   & $128 \times 128$ \\ \midrule
Normalisation $\beta$ & 128       & 256          & 128         & 128       \\
Time window         & 100 ms    & 50 ms        & 100 ms      & 100 ms    \\ \bottomrule
\end{tabular}%
}
\label{table:datasets}
\end{table}

\subsubsection{Model details}
\label{subsec:model}
{For the software implementation of the models, the Torch framework \cite{paszke2017automatic} was used along with the PyTorch Geometric library \cite{Fey2019PyTorchGeometric}. All experiments were performed on our three EFGCN models: Small, Base, and Large. Each model consists of 5 convolutional layers and two MaxPool layers. A~PoolOut layer was used at the end of each model, which is a~simple 2D MaxPool and a~single MLP layer for classification. Layer dimensions and exact configuration for all three variants are provided in the supplementary material.}

{Each model was implemented on the heterogeneous ZCU104 platform in two configurations: \( \beta \) $= 128$ with \texttt{TIME\_WINDOW} $=100$ ms and \( \beta \) $= 256$ with \texttt{TIME\_WINDOW} $=50$ ms.}

For a~200 MHz clock, each configuration meets the timing requirements.
Simulation results confirmed the consistency of the hardware module with the software model and the previously estimated throughput of 13.3 MEPS.
{For the \( \beta \) $= 128$ configuration, computing all output feature map elements independently for each convolution (i.e. with $m=1$) satisfies the required throughput.
However, for larger graphs and smaller time windows, parallelisation of certain multiplications is necessary.
The number of parallel vector multipliers required for each configuration is presented in Table \ref{table:multipliers} (details are available in the supplementary material).}

\begin{table}[!t]
\centering
\caption{{The number of parallel vector multipliers $m$ required to meet the throughput requirements for each model (with \( \beta \) $= 256$ and \texttt{TIME\_WINDOW} of 50 ms).}}
\resizebox{\columnwidth}{!}{%
\begin{tabular}{@{}ccccc|cccc@{}}
\toprule
 & \multicolumn{4}{c}{Output $dim$} & \multicolumn{4}{c}{Parallel multipliers $m$} \\
Model & Conv2 & Conv3 & Conv4 & Conv5 & Conv2 & Conv3 & Conv4 & Conv5 \\
\midrule
EFGCN-S & 32 & 32 & 32 & 32 & 8 & 8 & 1 & 1 \\
EFGCN-B & 32 & 32 & 64 & 64 & 8 & 8 & 2 & 2 \\
EFGCN-L & 32 & 64 & 64 & 128 & 8 & 16 & 2 & 4 \\
\bottomrule
\end{tabular}%
}
\label{table:multipliers}
\end{table}

\subsubsection{Training details}
In all experiments, the Adam optimiser \cite{kingma2014adam} was used along with the ExponentialLR scheduler. Each model was trained for 100 epochs, and for model quantisation the quantisation aware training (QAT) technique was applied for additional 50 epochs. During training, we used data augmentation in the form of event flipping along the X-axis (disabled for the MNIST-DVS dataset due to digit asymmetry) and \changed{spatial rotation of events in the (x, y) plane by a~uniformly sampled random angle around the event cloud centroid, with timestamps unchanged.}


\subsection{{Ablation studies}}
\label{ssec:ablation}

{Our ablation studies focus on three key aspects of the design: (i) model size, measured as the number of parameters (which directly affects on-chip memory usage), (ii) computational cost, measured in FLOPs (which is a~proxy for required throughput, latency, and energy per inference), and (iii) task performance, measured as classification accuracy. Unless stated otherwise, all ablations are performed on the Base model using the N-Cars dataset. For comparison, we include two recent event-based GCN approaches: EvGNN \cite{yanh_evgnn}, which provides a~hardware implementation of GCNs for event cameras without any pooling layers, and AEGNN \cite{schaefer2022aegnn}, which inspired both EvGNN and this work and incorporates 2D pooling.}

\subsubsection{{Impact of MaxPool modifications}}
\label{ssec:abl_pool}

{We start our analysis with the impact on the quality results of the modifications we applied to the pooling layers, which are shown in Table \ref{table:ablation_pool}.
In AEGNN, a~2D pooling method with an \textit{average} position was introduced, while in EvGNN the pooling layer was entirely omitted. Our results show that excluding the MaxPool significantly degrades performance, primarily due to its role in data generalisation in subsequent stages of the model, leading to improved results. Additionally, we observed that adopting a~3D pooling operation enhances accuracy by approximately 0.3-0.4\% compared to the 2D pooling method. In contrast, the use of \textit{divide} position has a~slight negative impact, reducing accuracy by around 0.1-0.2\%.}

\begin{table}[t!]
\caption{{Ablation studies on MaxPool modifications on accuracy on the N-Cars dataset show that the incorporation of a~3D MaxPool enhances overall performance. However, dividing the position component has been observed to slightly degrade the results. Bold indicates the configuration used in the hardware implementation.}
}
\centering
\label{table:ablation_pool}
\resizebox{0.7\columnwidth}{!}{\begin{tabular}{@{}ccccc@{}}
\toprule
2D MaxPool & 3D MaxPool & Average $\mathcal{P}$  & Divide $\mathcal{P}$ & ~~~~~~~Accuracy~~~~~~~ \\ \midrule
& & & & 0.887 \\
\checkmark &   & \checkmark         &   & 0.918 \\
\checkmark &  &  &  \checkmark & 0.917 \\
 &  \checkmark &  \checkmark  &  & 0.922 \\
  & \checkmark &  & \checkmark & \textbf{0.920} \\ \bottomrule
\end{tabular}
}
\end{table}

{Next, we compared the impact of using MaxPool layers and their absence in terms of the graph size and required operations.
Figure \ref{fig:flops_nodes_reduction} illustrates a~comparison of the total number of FLOPs performed by individual convolutions and the reduction in the number of vertices and edges in the graph. It can be observed that the number of FLOPs is closely correlated with the number of edges, primarily due to the operation of the PointNetConv layer, which applies a~linear transformation to each connection. Without the pooling operation, neither the number of edges nor vertices decreases, leading to 50 times more operations in subsequent convolutions, showing a~major drawback of the EvGNN solution.}

{These findings underscore the critical role of MaxPool layers, not only for improving qualitative results, but also for enhancing computational efficiency.}

\begin{figure}[t]
    \centering
    \begin{minipage}[t]{0.49\linewidth}
        \centering
        \begin{tikzpicture}
            \begin{axis}[
                tick label style={font=\scriptsize},
                label style={font=\scriptsize},
                legend style={font=\scriptsize},
                ybar,
                bar width=6pt,
                width=1.3\linewidth,
                height=5cm,
                ylabel={Total MFLOPs},
                xticklabel style={rotate=45, anchor=east},
                symbolic x coords={conv1,conv2,conv3,conv4,conv5},
                xtick=data,
                ymin=0, ymax=135,
                enlarge x limits=0.2,
                legend columns=-1,
                legend style={at={(0.5,1.05)}, anchor=south},
                grid=major,
                nodes near coords,
                nodes near coords align={vertical},
                every node near coord/.append style={font=\tiny},
            ]
                \addplot+[blue, fill=blue, opacity=0.7, every node near coord/.append style={xshift=-2pt}] coordinates {
                    (conv1,3.179189775)
                    (conv2,1.225806354)
                    (conv3,2.164814354)
                    (conv4,1.181742545) 
                    (conv5,2.213934545)
                };

                \addplot+[orange, fill=orange, opacity=0.7, every node near coord/.append style={xshift=2pt}] coordinates {
                    (conv1,3.179189775) 
                    (conv2,18.13373955) 
                    (conv3,30.69412355) 
                    (conv4,61.3882471) 
                    (conv5,111.6297831)
                };

                \legend{w. pooling, w/o pooling}

                \coordinate (A) at (axis cs:conv5,19.3882471);
                \coordinate (B) at (axis cs:conv5,111.6297831);

                \draw[<->, thick, gray]
                    ([xshift=-3pt]A) -- ([xshift=-3pt]B)
                    node[midway, xshift=-10pt, yshift=15pt, red, font=\scriptsize,inner sep=1pt]{x50!};
            \end{axis}
        \end{tikzpicture}
    \end{minipage}
    \hfill
    \begin{minipage}[t]{0.49\linewidth}
        \centering
        \begin{tikzpicture}
            \begin{axis}[
                tick label style={font=\scriptsize},
                label style={font=\scriptsize},
                legend style={font=\scriptsize},
                ybar,
                bar width=6pt,
                width=0.8\linewidth,
                height=5cm,
                ylabel={Reduction to input},
                xticklabel style={rotate=45, anchor=east},
                symbolic x coords={input,pool1,pool2},
                xtick=data,
                ymin=0, ymax=59,
                enlarge x limits=0.6,
                legend columns=-1,
                legend style={at={(0.5,1.05)}, anchor=south},
                grid=major,
                nodes near coords,
                nodes near coords align={vertical},
                every node near coord/.append style={font=\tiny},
            ]
                \addplot+[olive, fill=olive, opacity=0.7, every node near coord/.append style={xshift=-2pt}] coordinates {
                    (pool1, 6.15)
                    (pool2, 19.43)
                };

                \addplot+[teal, fill=teal, opacity=0.7, every node near coord/.append style={xshift=2pt}] coordinates {
                    (pool1, 13.38)
                    (pool2, 48.67)
                };

                \legend{vertices, edges}
            \end{axis}
        \end{tikzpicture}
    \end{minipage}
    \caption{{Comparison of the number of operations per second (left) and the number of vertices and edges (right) with and without the use of MaxPool layers. The use of pooling significantly reduces both the number of operations and graph complexity.}}
    \label{fig:flops_nodes_reduction}
\end{figure}

\subsubsection{Impact of all modifications}
\label{ssec:ablation_conv}

{To evaluate the impact of our convolution and graph generation modules along with the previously described MaxPool layers, we compared them with two state-of-the-art models in terms of size and performance. The AEGNN model consists of 7 SplineConv layers \cite{fey2018splinecnn} with one 2D MaxPool layer and two skip-connections, while the EvGNN model consists of 4 convolutional layers without any MaxPool layer.}

{To gain more comprehensive understanding of how our methods influence model size and performance, we introduced modified versions of both architectures: O-AEGNN and O-EvGNN. In these models, we replaced the original graph generation, convolutional, and MaxPool layers with our custom modules. Additionally, in the O-EvGNN model, we incorporated two MaxPool layers placed after the first and third convolutional layers.}

{The results presented in Table \ref{table:compare} demonstrate the impact of the proposed modifications. For the AEGNN, replacing SplineConv with our PointNetConv aggregation and adding our pooling strategy reduces the number of parameters from 30.4k to 5.0k, while slightly improving accuracy. This reduction is mainly due to the relatively high parameter cost of SplineConv layers \cite{jeziorek2023memory}. For the EvGNN, adding our pooling and graph modules yields accuracy gains, while also slightly reducing the number of parameters. In both cases, the modified versions (O-AEGNN and O-EvGNN) are more compact and more accurate than their original counterparts, demonstrating that our design improves both hardware efficiency and task performance.}

\begin{table}[]
\caption{
{Comparison of our methods with other architectures in terms of the number of parameters and accuracy for the N-Cars dataset. The results demonstrate that our solution has a~positive impact in both aspects.}}
\centering
\label{table:compare}
\resizebox{0.7\columnwidth}{!}{%
{
\begin{tabular}{@{}ccc@{}}
\toprule
~~~~~~Model~~~~~~   & ~~~~Accuracy~~~~ & ~~~~Num. of parameters~~~~  \\ \midrule
AEGNN   & 0.893    & 30.4k                \\
EvGNN   & 0.880    & 4.8k                  \\ \midrule
\textbf{O-AEGNN} & 0.898    & 5.0k                 \\
\textbf{O-EvGNN} & 0.891    & 4.2k                  \\ \bottomrule
\end{tabular}}}
\end{table}

\subsubsection{{Quantisation precision}}

{Although our hardware implementation was designed for 8-bit precision, it is crucial to explore the potential for quantising to lower bit precisions with QAT. In Figure  \ref{fig:quant_conv}, we compared the qualitative results and model sizes for different bit-widths. We analysed standard 8-bit, 7-bit, and 6-bit quantisations, and experimented with mixed precision quantisation, where part of the feature extraction was quantised to lower bit-widths (6/4 bits) while keeping the classifier at 8 bits.}

{The results show that 6-bit quantisation, which reduces the model size by approximately 23\%, has a~small impact on model quality,  while in the case of 4-bit quantisation there is a~considerable degradation in performance. It is worth noting that using 6-bit feature extractor with 8-bit classifier gives better results and size reduction of about 8\% compared to full 7-bit quantisation. In our models, where feature extraction part is implemented in the PL and the classifier in the PS, this modification could be crucial for deploying complex models.}

\begin{figure}[t!]
    \centering

    \begin{tikzpicture}
        \begin{axis}[%
            hide axis,
            xmin=0, xmax=1,
            ymin=0, ymax=1,
            legend style={
                legend columns=3,
                font=\scriptsize,
                /tikz/every even column/.append style={column sep=0.1cm},
                anchor=center
            }
        ]
        \addlegendimage{mark=*, mark size=1, color=red, solid};
        \addlegendentry{Small};
        \addlegendimage{mark=*, mark size=1, color=blue, solid};
        \addlegendentry{Base};
        \addlegendimage{mark=*, mark size=1, color=black, solid};
        \addlegendentry{Large};
        \end{axis}
    \end{tikzpicture}

    \vspace{.5em}

    \begin{subfigure}[t]{0.48\linewidth}
        \centering
        \begin{tikzpicture}
            \begin{axis}[
                tick label style={font=\scriptsize},
                height=5cm,
                label style={font=\scriptsize},
                symbolic x coords={8-bit, 7-bit, 6+8-bit, 6-bit, 4+8-bit},
                xtick=data,
                xticklabel style={rotate=45, anchor=east},
                ylabel={Accuracy [\%]},
                ymin=85, ymax=93,
                width=7cm,
                grid=major,
                cycle list name=color list
            ]
                \addplot+[mark=*, mark size=1] coordinates {(8-bit, 91) (7-bit, 90.45) (6+8-bit, 90.7) (6-bit, 90.21) (4+8-bit, 85.38)};
                \addplot+[mark=*, mark size=1] coordinates {(8-bit, 91.87) (7-bit, 91.43) (6+8-bit, 91.67) (6-bit, 91.18) (4+8-bit, 87.79)};
                \addplot+[mark=*, mark size=1] coordinates {(8-bit, 92.49) (7-bit, 92.25) (6+8-bit, 92.33) (6-bit, 92.08) (4+8-bit, 86.89)};
            \end{axis}
        \end{tikzpicture}
        \caption{Accuracy}
    \end{subfigure}
    \begin{subfigure}[t]{0.48\linewidth}
        \centering
        \begin{tikzpicture}
            \begin{axis}[
                tick label style={font=\scriptsize},
                height=5cm,
                label style={font=\scriptsize},
                symbolic x coords={8-bit, 7-bit, 6+8-bit, 6-bit, 4+8-bit},
                xtick=data,
                xticklabel style={rotate=45, anchor=east},
                ylabel={Size [kB]},
                width=7cm,
                grid=major,
                cycle list name=color list
            ]
                \addplot+[mark=*, mark size=1] coordinates {(8-bit, 5.51) (7-bit, 4.89) (6+8-bit, 4.523) (6-bit, 4.27) (4+8-bit, 3.54)};
                \addplot+[mark=*, mark size=1] coordinates {(8-bit, 10.94) (7-bit, 9.68) (6+8-bit, 8.91) (6-bit, 8.41) (4+8-bit, 6.88)};
                \addplot+[mark=*, mark size=1] coordinates {(8-bit, 20.6) (7-bit, 18.18) (6+8-bit, 16.75) (6-bit, 15.75) (4+8-bit, 12.9)};
            \end{axis}
        \end{tikzpicture}
        \caption{Size}
    \end{subfigure}
    
    \caption{Comparison of models accuracy and size for different precisions \changed{on the N-Cars dataset}. The results indicate that quantisation to 6 bits still enables high accuracy for each model.}
    \label{fig:quant_conv}
\end{figure}

\subsection{Comparison with other works}

In this Section, we compare our three models (Small, Base and Large) with the state-of-the-art based on the N-Cars, N-Caltech101, CIFAR10-DVS, and MNIST-DVS datasets. The results in terms of accuracy, FLOPs per event, and network size are presented in Table \ref{table:sota_acc}. We primarily selected methods that use graph-based event representations \cite{bi2019graph, li2021graph, schaefer2022aegnn, yanh_evgnn}, but also included other asynchronous methods, such as SNNs \cite{orchard2015hfirst} and dense neural networks \cite{cannici2019asynchronous, messikommer2020event}. \changed{Although FLOPs per event is a~widely used normalised proxy for computational complexity in software-oriented comparisons, and provides a~useful indication of the overall complexity trend of a~given model, it does not fully characterise implementation cost on FPGA platforms, where factors such as numerical precision, memory access patterns, and data movement also play a~major role. For this reason, the comparison in Table \ref{table:sota_acc} should be interpreted together with the hardware-level resource utilisation and latency analysis reported in Table \ref{table:utils_acc}.}



\changed{A~few conclusions can be drawn from the results. First, our models achieve the lowest FLOPs per event among the compared methods on all datasets for which this metric is available. The difference is particularly large with respect to dense asynchronous approaches such as YOLE and AsyNet, reaching more than three orders of magnitude in some cases. Compared with prior graph-based methods, our EFGCN also maintains a~favourable complexity profile; for example, EvGNN, despite using fewer parameters than our Base model, still requires a~higher number of FLOPs per event. This indicates that the proposed 3D MaxPool-based design provides a~substantial computational advantage. Second, our models are among the smallest in terms of the number of parameters, with only EvGNN being smaller. At the same time, the substantially larger models, such as AEGNN, do not translate into a~proportionally better efficiency-accuracy trade-off; for the N-Caltech101, the gap in parameter count reaches up to two orders of magnitude.}

\changed{In terms of accuracy, our Small model outperforms G-CNNs, HATS, EvGNN, and YOLE on individual datasets. Our Large model performs better than NvS-S, while remaining close to YOLE on the N-Cars, AEGNN and HATS on the N-Caltech101, and EvS-S on the MNIST-DVS. It should also be noted that some graph-based methods use relatively large edge-generation radius (e.g. 5 in AEGNN for the N-Caltech101 and 5 in NvS-S/EvS-S for all datasets), which has a~strong impact on accuracy, as analysed in detail in the supplementary material.}

\begin{table*}[]
\centering
\caption{Comparison with other methods of object classification {for Float32 models. The results show that our models are among the smallest in terms of the number of parameters and exhibit the lowest computational complexity per event, while achieving results comparable to the state-of-the-art. The values obtained from the open source code are in parentheses.}}
\resizebox{0.99\textwidth}{!}{%
\begin{tabular}{@{}cccccccccccc@{}}
\toprule
\textbf{} &
  \textbf{} &
  \textbf{} &
  \multicolumn{2}{c}{N-Cars} &
  \multicolumn{2}{c}{N-Caltech101} &
  \multicolumn{2}{c}{CIFAR10-DVS} &
  \multicolumn{2}{c}{MNIST-DVS} &
  \textbf{} \\ \cmidrule(lr){4-11}
Model    & Repr.        & Async.                    & Acc. $\uparrow$  & MFLOPs/ev $\downarrow$ & Acc. $\uparrow$  & MFLOPs/ev $\downarrow$ & Acc. $\uparrow$  & MFLOPs/ev $\downarrow$ & Acc. $\uparrow$   & MFLOPs/ev $\downarrow$ & \# Param. $\downarrow$      \\ \midrule
{H-First \cite{orchard2015hfirst}}  & {Spike}        & \checkmark & {0.561} & -         & {0.054} & -         & -     & -         & -     & -         & -              \\
{HATS \cite{sironi2018hats}}    & {Time-Surface} & \checkmark & {0.902} & {0.03}      & {0.642} & {4.3}       & {0.524} & {0.18}      & {0.984} & {0.18}      & -              \\
{YOLE \cite{cannici2019asynchronous}}    & {Voxel-Grid}   & \checkmark & {0.927} & {328.16}    & {0.702} & {3659}      & -     & -         & {0.961} & -         & {2.03 M}         \\
{AsyNet \cite{messikommer2020event}}  & {Voxel-Grid}   & \checkmark & {0.944} & {21.5}      & {0.745} & {202}       & {0.663} & {103}       & {0.994} & {112}       & {3.69 M}         \\ \midrule
G-CNNs \cite{bi2019graph}  & Graph        & \xmark     & 0.902 & -         & 0.630 & -         & 0.515 & -         & 0.974 & -         & 4.93 M         \\
RG-CNNs \cite{bi2019graph}  & Graph        & \xmark     & 0.914 & -         & 0.657 & -         & 0.540 & -         & 0.986 & -         & 5.10 M         \\
NvS-S  \cite{li2021graph}   & Graph        & \checkmark & 0.915 & 5.2       & 0.670 & 7.8       & 0.602 & 22.8      & 0.986 & 10.1      & 0.36 - 1.16 M  \\
EvS-S  \cite{li2021graph}   & Graph        & \checkmark & 0.931 & 6.1       & 0.761 & 11.5      & 0.680 & 33.2      & 0.991 & 15.2      & 0.36 - 1.16 M  \\
AEGNN  \cite{schaefer2022aegnn}  & Graph        & \checkmark & 0.945 (0.893) & 0.47      & 0.668 (0.643) & 7.31      & -     & -         & -     & -         & 0.03 - 20.4 M  \\
{EvGNN \cite{yanh_evgnn}}   & {Graph}        & \checkmark & {0.880} & {0.063}         & -     & -         & -     & -         & -     & -         & {4.8 k}          \\ \midrule
{\textbf{EFGCN-S}}  & {Graph}        & \checkmark & {0.907} & {0.032}    & {0.623} & {0.004}    & {0.571} & {0.011}    & {0.964} & {0.006}    & {5.2 - 55.5 k}   \\
{\textbf{EFGCN-B}}  & {Graph}        & \checkmark & {0.920} & {0.060}    & {0.631} & {0.008}    & {0.588} & {0.022}    & {0.978} & {0.012}    & {10.6 - 111.0 k} \\
{\textbf{EFGCN-L}}  & {Graph}        & \checkmark & {0.923} & {0.111}    & {0.639} & {0.015}    & {0.613} & {0.040}    & {0.983} & {0.022}    & {20.2 - 220.9 k} \\ \bottomrule
\end{tabular}
\label{table:sota_acc}
}
\end{table*}

\subsection{Evaluation of hardware implementation}
\label{ssec:HW-eval}

{In Table \ref{table:utils_acc}, we present a~comparison with the solutions proposed in the literature on event-based object classification systems implemented in FPGAs in terms of resource utilisation, model accuracy after quantisation and latency.
\begin{table*}[]
\centering
\caption{{Comparison of the \ours\ with the SOTA in terms of resource utilisation, accuracy and latency on SoC FPGAs.}}
\resizebox{0.99\textwidth}{!}{%
{\begin{tabular}{@{}cccccccc|cccc|ccc@{}}
\toprule
\textbf{} &
  \textbf{} &
  \textbf{} &
  \multicolumn{5}{c}{Utilisation} &
  \multicolumn{4}{c}{Accuracy [\%]} &
  \multicolumn{3}{c}{Latency} \\ 
\cmidrule(lr){4-15}
Model & Graph size & Time window & LUT & FF & BlockRAM & UltraRAM & DSP & N-Cars & N-Caltech101 & CIFAR & MNIST & PL [ms] & PL+PS [ms] & per ev. [$\mu$ s]
\\ \midrule

\textbf{EFGCN-S} & 128 & 100 ms & 43553 & 15292 & 121.5 & 0 & 90 & 91.0 & - & 57.4 & 98.7 & 3.78 & 4.39 & 6.56 \\ 
\textbf{EFGCN-B} & 128 & 100 ms & 53517 & 18798 & 162 & 0 & 90 & 91.9 & - & 59.1 & 97.9 & 4.62 & 5.77 & 9.44 \\ 
\textbf{EFGCN-L} & 128 & 100 ms & 65901 & 24622 & 211.5 & 0 & 90 & 92.5 & - & 61.5 & 98.5 & 7.03 & 9.31 & 13.76 \\ 
\midrule
\textbf{EFGCN-S} & 256 & 50 ms & 112670 & 24087 & 149 & 12 & 172 & - & 62.8 & - & - & 4.42 & 5.12 & 4.02 \\ 
\textbf{EFGCN-B} & 256 & 50 ms  & 139382 & 29401 & 187.5 & 12 & 184 & - & 63.2 & - & - & 4.43 & 5.77 & 4.04 \\ 
\textbf{EFGCN-L} & 256 & 50 ms  & 142773 & 38882 & 240 & 12 & 1364 & - & 64.1 & - & - & 4.43 & 7.11 & 4.04 \\ 
\midrule
\textbf{EvGNN} \cite{yanh_evgnn} & $120 \times 100$ & - & 30908 & 24083 & 15 & 48 & 228 & 87.8 & - & - & - & - & - & ~16\\
\textbf{ESDA} \cite{gao2024composable} & $180 \times 240$ & - & 154000 & 115000 & 1278 & 0 & 1792 & - & 72.4 & - & - & 3.09 & - & -\\
\bottomrule
\end{tabular}}
\label{table:utils_acc}
}
\end{table*}

We implemented each of the EFGCN network models (Small, Base, Large) for hardware testing and evaluation. Our design was presented in the form of a~demonstration during 2025 CVPR conference \cite{wzorek2025live}.

\subsubsection{Resource utilisation}

As presented in Table \ref{table:utils_acc}, our design demonstrates high scalability and adaptability across different model sizes and input resolutions.
As the graph size and feature dimensionality increase, both logic and memory utilisation naturally grow.
However, this can be controlled by adjusting the number of parallel multiplications according to the available hardware resources and throughput requirements.
In this work, we assume the worst-case scenario regarding data sparsity -- that is, the throughput is designed to handle all possible nodes with all possible connections.
The prominent increase in logic utilisation observed for models implemented with the \( \beta \) $= 256$ results from the larger number of parallel multipliers.
\changed{A~detailed analysis of the impact of individual system components and successive hardware modules on the overall system utilisation is presented in the supplementary material.}

DSP utilisation also depends on the number of parallel multipliers, as they are required for requantisation.
For Large model with the \( \beta \) $= 256$ configuration, part of the multiplication operations was implemented using DSPs instead of LUTs to reduce LUT usage, which explains the variable DSP utilisation.

\changed{The direct impact of the number of multipliers on resource utilisation is presented in Table \ref{table:scalability}, using the Small model with \( \beta \) $= 128$ as an example. Increasing the number of parallel multipliers directly translates into higher usage of logic resources (used for multiplication) and DSP blocks (used for requantisation). The number of parallel multipliers does not affect memory consumption in the system, since both the number of weights and the stored feature maps remain unchanged.}

While increasing graph sizes pose a~challenge for our method, as they lead to higher resource utilisation, future work will focus on processing dynamically updated voxel graphs, where each vertex represents events generated by multiple pixels. This approach should enable efficient processing of data captured by higher-resolution sensors without the need to increase the graph size.

\changed{Table \ref{table:scalability} also presents system utilisation with quantisation applied to both weights and feature maps reduced to 6 bits instead of 8 bits. Reducing numerical precision leads to lower memory consumption as well as reduced usage of LUTs/FFs (for feature vector buffering and vector multiplication). DSP utilisation remains unchanged, as DSP blocks are used exclusively for requantisation performed on 32-bit values.}

\begin{table*}[]
\centering
\caption{\changed{Experimental results demonstrating the scalability of \ours\, illustrated using the Small model (graph size 128), for varying numbers of vector multipliers (m) and quantisation to 8 and 6 bits.}}
\resizebox{0.99\textwidth}{!}{%
{\begin{tabular}{@{}ccccc|ccc|cc@{}}
\toprule
\textbf{} &
  \multicolumn{4}{c}{Utilisation} &
  \multicolumn{3}{c}{Accuracy [\%]} &
  \multicolumn{2}{c}{Latency} \\ 
\cmidrule(lr){2-10}
Model & LUT & FF & BlockRAM & DSP & N-Cars & CIFAR & MNIST & PL [ms] & per ev. [$\mu$ s]
\\ \midrule

\textbf{8-bit, m = 1} & 41350 & 12960 & 121.5 & 88 & 91.0 & 57.4 & 98.7 & 3.78 & 6.56 \\
\textbf{6-bit, m = 1} & 31332 & 10886 & 95 & 88 & 90.7 & 57.2 & 98.5 & 3.78 & 6.56 \\
\textbf{8-bit, m = 2} & 66001 & 16011 & 121.5 & 112 & 91.0 & 57.4 & 98.7 & 1.93 & 3.66 \\ 
\textbf{8-bit, m = 4} & 114513 & 21608 & 121.5 & 160 & 91.0 & 57.4 & 98.7 & 1.02 & 2.21 \\
\bottomrule
\end{tabular}}
\label{table:scalability}
}
\end{table*}

\subsubsection{Latency}

The latency of the PL (graph generation, feature extraction) is defined as the time between the registration of the last event in a~given sequence and the reception of the final feature map in the PS.
The total latency, including additional processing of the network's head, is defined as PL+PS latency.
Processing times in the PL part were measured using the ILA (Integrated Logic Analyser) tool, while the total latency (with the PS) was measured with timing functions available in the Vitis software.

The proposed EFGCN model enables predictions to be generated every 25 ms with latency of 4.39-9.31 ms (for the configuration with \texttt{TIME\_WINDOW} $=100$ ms), or every 12.5 ms with a~latency of 5.12-7.11 ms (for the configuration with \texttt{TIME\_WINDOW} $=50$ ms). 
Moreover, the use of elements with fixed latency only (no access to external memory for convolutional layers) allowed for the determination of a~constant system throughput of 13.3 MEPS, the value not achievable for alternative solutions.

The analysis of the achieved latency leads to some interesting conclusions.
Firstly, for models with \texttt{TIME\_WINDOW} $=100$ ms configuration, increasing the model size leads to higher latency, because for each convolution we set a~number of parallel multipliers $m=1$. 
As the feature map sizes increase, the number of necessary multiplications grows, and consequently, the time required to perform them.

This issue does not occur in \texttt{TIME\_WINDOW} $=50$ ms configuration. 
Regardless of the feature map sizes, the measured latency remains similar -- close to 4.43 ms. 
With more stringent throughput requirements, different multiplication strategies were adopted for each layer.
The larger the number of elements in the feature map, the more parallel multipliers were used.
As a~result, the amount of utilised resources increases, but the latency remains constant.

Based on these observations, it can be concluded that in the proposed system latency and resource usage are correlated.
\changed{This assumption was confirmed by the experiment documented in Table \ref{table:scalability}. As the number of parallel multipliers increases, the system latency decreases. By relying exclusively on internal memory resources and providing direct low-latency access to the corresponding features, the system latency for a~given configuration of graph convolution modules becomes predictable.
Appropriate architectural configuration should be thus motivated by the specific requirements set for the given application.}

\subsubsection{Power consumption}

Using the Vivado Power Analysis tool, we estimated that the total power consumption of the programmable logic for each configuration ranges from 1.83 W (Small model, $\beta = 128$) to 5.48 W (Large model, $\beta = 256$).
It should be noted, however, that the actual power usage strongly depends on input toggle rates, while event data is inherently sparse across all dimensions.
Based on this observation, we can assume that the real power consumption associated with processing actual event data on hardware would be lower.

To validate this assumption, we performed a~detailed power analysis in Vivado for the Base model, using input toggle rates estimated from post-implementation simulations on a~representative sample from the DVS-MNIST/N-Caltech101 datasets (achieving 99.9\% of annotated nets).
Under these conditions, the estimated power consumption of the PL decreased from 2.6 W (baseline estimation for $\beta = 128$) to 1.23 W, and from 5.34 W (baseline estimation for $\beta = 256$) to 1.86 W.

For future work, we plan to conduct a~comprehensive study of the actual power consumption of the EFGCN implemented on the ZCU104 platform, as well as its dependence on scene dynamics (i.e. the number of events per second).

The estimated power used by the PS part (ARM processor) is 2.64 W, regardless of the selected configuration and model. 
Nevertheless, the network’s head could easily be integrated into the programmable logic, which would considerably reduce latency and power consumption with only a~minor impact on resource utilisation. The final linear layers could be computed with limited parallelism, as this operation is performed infrequently -- once every 25 ms (for larger graphs) or 12.5 ms.

\subsubsection{Comparison with other hardware implementations}
\label{ssec:hw_sota_compare}

With the limited number of experiments conducted by alternative solutions, direct comparison is possible using only the N-Cars and the N-Caltech101 datasets.

For the \textbf{N-Cars} dataset, our method achieves the highest accuracy scores among all those proposed in the literature for each GCN model (Small, Base, and Large). 
Moreover, our Small model achieves a~3.2\% higher accuracy than EvGNN \cite{yanh_evgnn}, with 41\% higher LUT utilisation, while simultaneously reducing FF resources by 36\%, DSPs by 60\%, and internal memory resources by 70\% (estimation based on a~32 kb capacity for BlockRAM and 288 kb for UltraRAM).
Therefore, our Small model achieves higher accuracy and lower value of MFLOPs/ev than the EvGNN model.
In the EvGNN paper \cite{yanh_evgnn}, the system latency for entire sequences was not measured, and instead, latency per event was used as the metric.
To enable a~direct comparison of these methods, Table \ref{table:utils_acc} also presents the per-event latency calculated as the sum of the delays of the individual hardware modules.
However, it should be noted that this metric is not perfect for our method, as it does not reflect its operational nature (primarily due to the synchronisation of computations and feature map buffering induced by the MaxPool layers -- Section \ref{sec:FPGA}).

\changed{While some of the hardware-aware optimisations proposed in our previous work (the use of a~directed graph \cite{jeziorek2024optimising} and PointNetConv \cite{jeziorek2023memory}) are applied both in the proposed system \ours\ and in EvGNN, the key difference between those methods is the three-dimensional MaxPool mechanism employed in our approach, which enables exploitation of the temporal sparsity of the data and improved scalability achieved through sequential multiplications in the PointNetConv layers.}
The system described in the former achieves low latency by processing events asynchronously for the entire network.
However, it should be emphasised that in doing so, the authors opted out of data scaling between convolutional layers, which leads to a~significant increase in computational complexity (Section \ref{sec:FPGA}). 
To maintain low latency in the system, feature maps of small sizes (maximum of 32 elements) were used, and a~matrix multiplication strategy based on a~high level of parallelism (MAC engine per output channel) was adopted.

By using 3D MaxPool, EFGCN offers better scalability through efficient computations, achieving higher classification accuracy with the use of fewer resources.
\changed{With the aggregation of vertices over time and the use of an on-chip memory with constant latency, it is possible to derive strict throughput requirements for subsequent graph convolutions and accordingly adapt the computational architecture. This also enables the use of higher feature dimensionality.}
The ability to configure the level of parallelism in matrix multiplication allows for adjusting the achieved latency to the specific computer vision problem.
In the case of EvGNN, on the other hand, increasing the size of individual feature maps or the number of layers impacts both latency and resource usage.

\changed{At the same time, with 3D MaxPool the number of features that must be stored in memory for each graph convolution module has also been reduced, as well as the number of required computations, through an effective reduction of the search radius from 3 to 1. This minimises the maximum number of matrix multiplications and memory accesses from 29 to 18.}

In summary, the EFGCN is more scalable for \changed{deeper models}, data with higher resolutions and greater dynamics.

It should also be emphasised that the EvGNN system utilises external DDR memory resources, which leads to variable latency and significantly increases power consumption.
\changed{In the literature, it is commonly assumed that access to external DRAM memory is 10–1000 $\times$ more energy-expensive than access to on-chip memory \cite{horowitz20141}. In \cite{yanh_evgnn}, it was shown that over 50\% of system latency and 73\% of energy consumption are attributable to DRAM communication. This type of overhead has been eliminated in \ours\ through the exclusive use of an on-chip memory.
It should be noted, however, that} for the EvGNN model, power consumption is reported only for the ASIC implementation, which makes a~direct comparison with our model infeasible.

In the case of the \textbf{N-Caltech101} dataset, our module allows for classification with accuracy ranging from 62.8\% to 64.1\%.
The ESDA system proposed in \cite{gao2024composable}, which utilises sparse convolutional neural networks, achieves higher accuracy. 
However, the resource utilisation of this method is higher across all considered elements.
The main issue is the very high usage of internal memory resources (between 5.3 and 8.5 times higher compared to the individual EFGCN models). 
Implementing ESDA for the ZCU104 platform considered in this paper would not be feasible due to the usage of FF, DSP, and BlockRAM resources exceeding the available number of elements (in the case of BlockRAMs -- 4 times).
At the same time, given the motivation related to the use of event-based cameras for mobile robotics, limiting resource utilisation is one of our priorities.

While the latency of 3.09 ms achieved in the ESDA model is nominally lower than the one achieved by EFGCN, it is important to consider the differences in operation.
Due to the use of two-dimensional event-frame representations, systems utilising convolutional neural networks require prior data buffering and frame generation.
In contrast, our system processes events directly as soon as they are registered.
EFGCN allows for predictions 4 times more frequently than the \texttt{TIME\_WINDOW} value, thus providing greater throughput than a~solution requiring prior data accumulation.

The achieved power consumption values are higher than those obtained in the ESDA model (1.81 W measured during runtime for FPGA fabric using the built-in power monitor controlled by the PMB). 
However, it is important to emphasise the differences in methodology (estimation vs measurement) as well as the fact that in the ESDA model the representation generation in the processor part of the system is necessary (introducing additional power requirements).
}

\subsection{Comparison with embedded GPU}

{To evaluate the time and energy efficiency of our method compared to alternative embedded platforms, a~reference implementation was created on the Jetson Orin NX eGPU platform. This system features an ARM Cortex-A78 processor and an NVIDIA Ampere GPU. Testing conditions similar to the FPGA configuration were reproduced by implementing a~graph generator in C++, operating on a~single CPU core. This generator reads events from the memory according to their timestamps, simulating the operation of a~real event camera.}

{The resulting graph was then processed in parallel by the model on both CUDA and CPU platforms at two power levels: 10 W and 25 W. Figure \ref{fig:jetson} presents the average total power consumption and latency for all datasets. Due to the small size of the tested models, differences in processing times between EFGCN variants were minimal; therefore, a~detailed analysis is provided only for the Base model.}

{Propagation times for models running on the CUDA platform ranged from 18.3 to 22.8 ms at 25 W and from 22.0 to 27.1 ms at 10 W for the CIFAR10-DVS, MNIST-DVS, and N-Cars datasets. Compared to the slowest hardware configuration (EFGCN-L with PL+PS mode), a~speed-up of 2.6–3.2 and 3.0–3.9 times for FPGA was achieved, respectively. Additionally, the average graph generation time was 0.57–0.63 ms for the CPU on the Jetson platform, while FPGA implementation required only 75 µs (15 clock cycles at a~200 MHz frequency). Considering this difference, the FPGA platform offers an overall speed-up of $7.6$–$8.4\times$ with comparable power consumption. Moreover, the average data transfer time from CPU to CUDA was 3.5 ms, introducing additional latency to the system.}

{Similar tests conducted exclusively on CPU-based models indicated even greater advantage of the FPGA implementation, achieving speed-ups of 18.9–\\127.8 times (25 W) and 20.4–139.9 times (10 W), with comparable graph generation times and without the need of data transfer to CUDA.}

{The presented results confirm the significant advantage of the proposed hardware implementation of graph convolutional networks over solutions utilising embedded GPU systems.}

\begin{figure}[t]
\centering
\begin{tikzpicture}
    \begin{axis}[
        tick label style={font=\scriptsize},
        label style={font=\scriptsize},
        legend style={font=\scriptsize},
        ybar=5pt,
        bar width=7pt,
        width=\linewidth,
        height=4cm,
        ymin=0, ymax=10,
        ylabel={Power [W]},
        symbolic x coords={CIFAR,MNIST,N-Cars,N-Caltech101},
        xtick=data,
        enlarge x limits=0.2,
        legend style={at={(0.5,1.4)}, anchor=north, legend columns=-1},
        xtick pos=left,
        grid=major,
        nodes near coords,
        nodes near coords align={vertical},,
        every node near coord/.append style={font=\tiny},
        ]
        \addplot coordinates {(CIFAR,6.482) (MNIST,6.144) (N-Cars,6.179) (N-Caltech101,6.755)};
        \addplot coordinates {(CIFAR,8.559) (MNIST,8.288) (N-Cars,8.353) (N-Caltech101,8.521)};
        \addplot coordinates {(CIFAR,6.090) (MNIST,5.764) (N-Cars,5.892) (N-Caltech101,6.430)};
        \addplot coordinates {(CIFAR,6.821) (MNIST,6.640) (N-Cars,6.718) (N-Caltech101,6.845)};
        \addplot coordinates {(CIFAR,3.87) (MNIST,3.87) (N-Cars,3.87) (N-Caltech101,4.50)};

        \legend{CUDA 25 W, CPU 25 W, CUDA 10 W, CPU 10 W, FPGA}
    \end{axis}
\end{tikzpicture}

\begin{tikzpicture}
    \begin{axis}[
        ymode=log,
        tick label style={font=\scriptsize},
        label style={font=\scriptsize},
        legend style={font=\scriptsize},
        ybar=5pt,
        bar width=7pt,
        width=\linewidth,
        height=4cm,
        ymin=0, ymax=5000, 
        ylabel={Latency [ms]},
        symbolic x coords={CIFAR,MNIST,N-Cars,N-Caltech101},
        xtick=data,
        enlarge x limits=0.2,
        legend style={at={(0.5,1.25)}, anchor=north, legend columns=-1},
        xtick pos=left,
        grid=major,
        nodes near coords align={vertical},
        every node near coord/.append style={font=\tiny},
        visualization depends on={rawy \as \rawy}, 
        nodes near coords style={
            /pgf/number format/fixed,
            /pgf/number format/precision=2,
        },
        nodes near coords*={\pgfmathprintnumber{\rawy}}, 
        ]
        \addplot coordinates {(CIFAR,22.8) (MNIST,18.3) (N-Cars,19.8) (N-Caltech101,25.9)};
        \addplot coordinates {(CIFAR,898.2) (MNIST,133.2) (N-Cars,265.8) (N-Caltech101,1201.9)};
        \addplot coordinates {(CIFAR,27.1) (MNIST,21.3) (N-Cars,22.0) (N-Caltech101,28.2)};
        \addplot coordinates {(CIFAR,983.6) (MNIST,143.2) (N-Cars,308.9) (N-Caltech101,1432.5)};
        \addplot coordinates {(CIFAR,5.77) (MNIST,5.77) (N-Cars,5.77) (N-Caltech101,5.77)};
    \end{axis}
\end{tikzpicture}
    \caption{{Power consumption and inference latency of the Base model on the Jetson Orin NX in two power modes (10~W and 25~W), using either the CPU or CUDA cores, compared with the proposed implementation for the ZCU104 SoC FPGA.}} 

\label{fig:jetson}
\end{figure}

\section{Limitations}
\label{sec:limitations}

{In this section we define the current limitations of our system and outline the methods planned to address them with the further research on applying the accelerator to more complex computer vision tasks (e.g. object detection).}

{To increase data resolution and allow smaller temporal windows, a~higher number of parallel multipliers is required.
Consequently, the consumption of logic resources would increase.
We plan to investigate techniques aimed at optimisation of multiplications (e.g. further utilisation of DSP resources with DSP packing).
Additionally, we plan to explore the application of pruning techniques for the considered GCN model.
Removing certain weights in the model would reduce both memory (sparse features) and logic resource usage (fewer multiplications required for a~given graph convolutional layer).}

{In the current implementation, we support only a~fixed graph edge search radius of $R=3$. We plan to evaluate the system with other values and conduct studies on their accuracy and efficiency.
For more complex computer vision tasks, skip-connections are a~widely used technique. We plan to integrate this mechanism by utilising external memory in a~way that preserves the system's fixed latency.
Furthermore, we are considering the implementation of a~voxel-graph approach, where representing multiple events with a~single node could significantly reduce the required memory and logic resources.}

{Finally, we plan to limit the use of the processing system in order to further reduce both power consumption and system latency. We aim to achieve this, by implementing the network’s classification head with the AI Engines specific to the AMD Versal platform, instead of relying on the PS.}

\section{Summary}
\label{sec:summary}

{In this work, we presented a~hardware implementation of a~graph convolutional neural network for event data processing on a~SoC FPGA platform.
The developed system enables continuous event data processing, generating predictions for the last 50/100 ms (depending on the configuration) while maintaining a~fourfold increase in output frequency.
Extensive evaluations across multiple models, configurations, and datasets demonstrate that the proposed approach outperforms existing methods in classification accuracy, computational efficiency, and hardware resource utilisation. 
Notably, our implementation operates without \changed{any off-chip} external memory, leading to reduced power consumption.
Moreover, we demonstrated the superior performance of the proposed hardware system compared to other platforms, namely the CPU and GPU.

Our solution surpasses state-of-the-art methods while offering high scalability. By leveraging the inherent temporal sparsity of event data and 3D MaxPool layers, we reduce FPGA logic utilisation through sequential computation of selected operations.
Furthermore, the system establishes a~direct trade-off between latency and resource utilisation -- allowing for configuration adjustments that optimise one metric at the expense of the other, depending on the application’s requirements.

As part of future work, we plan to introduce additional hardware optimisations, such as initial voxelisation of the graph, to reduce both representation size and computational complexity. Additionally, we aim to integrate techniques that enhance the training process, including pruning and skip-connections. The ultimate goal is to develop a~fully hardware-accelerated GCNN capable of real-time event-based object detection.
This work addresses the application of our method for event-based vision, but it can also be used for other types of spatio-temporal point cloud data.}

\section{Acknowledgements}
\label{sec:ack}
This work was supported by the ``Excellence initiative -- research university'' programme for AGH University of Kraków, the Polish National Science Centre projects 2024/53/N/ST6/04254 and 2024/53/N/ST6/04331, Polish high-performance computing infrastructure PLGrid (HPC Center: ACK Cyfronet AGH -- grant no. PLG/2023/016897) and 16.16.120.773 for the AGH University of Krakow.

\bibliographystyle{elsarticle-num} 
\bibliography{IEEEexample}

\newpage
\includegraphics[width=1in,height=1.25in,clip,keepaspectratio]{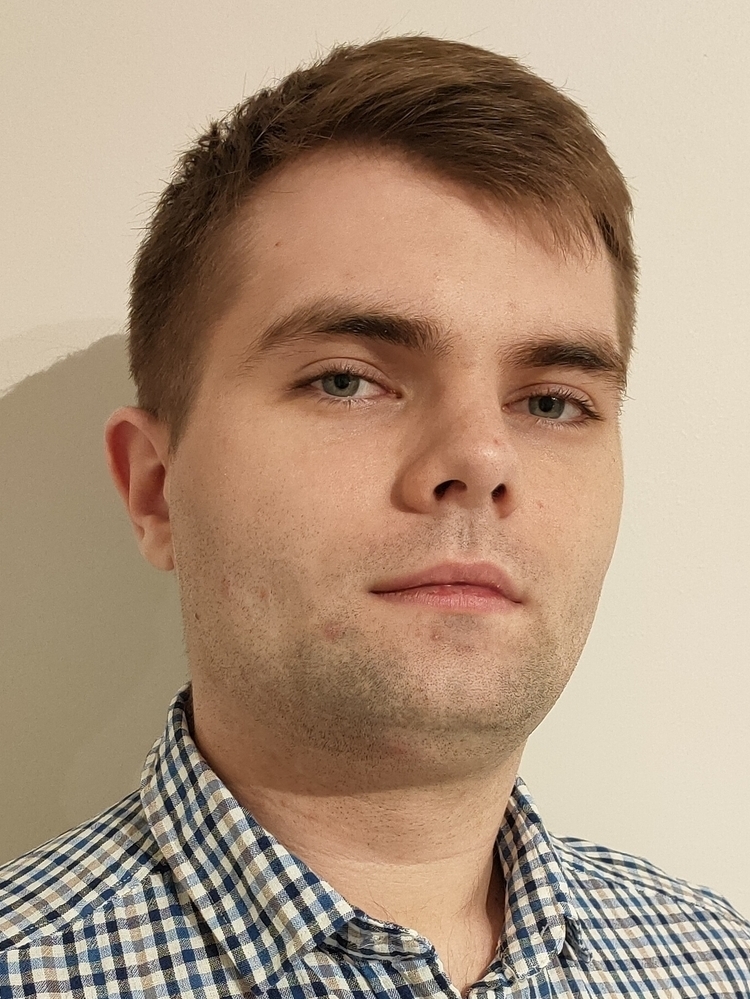}
{\textbf{Kamil Jeziorek} received the Master of Science degree in Automation and Robotics in 2023 at AGH University of Kraków, Poland. He is currently pursuing the Ph.D. degree at the Department of Automatics and Robotics, Laboratory of Vision Systems, under the supervision of Prof. dr hab. inż. Marek Gorgoń. His research interests include embedded vision systems based on event-driven cameras, graph neural networks, and hardware acceleration of deep learning pipelines on FPGA platforms.}

{\includegraphics[width=1in,height=1.25in,clip,keepaspectratio]{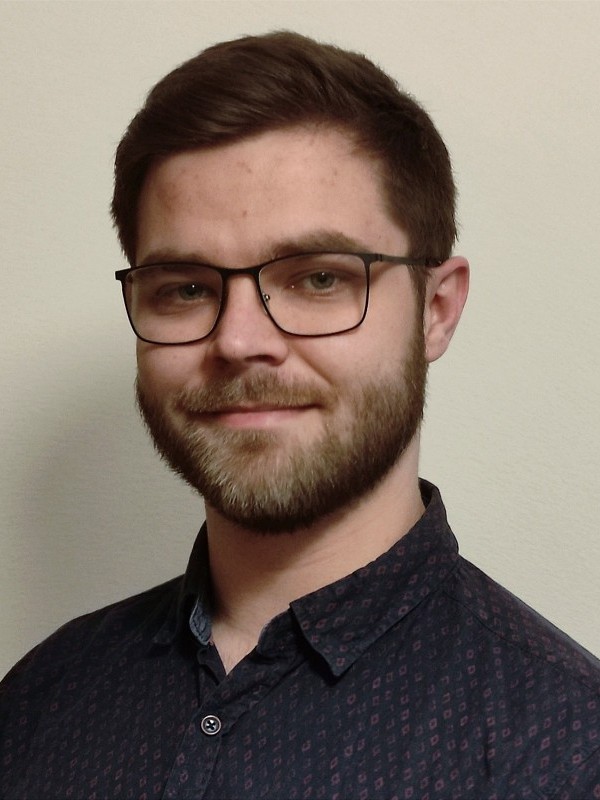}}
{\textbf{Piotr Wzorek} received his Engineering Degree in
Automation and Robotics in 2020 and his Master
of Science degree in Intelligent Control Systems in
2021. He is currently pursuing a Ph.D. degree and
works as a research assistant at AGH University of
Krakow.  His research interests include the application of neuromorphic sensors for mobile robotics, the acceleration of deep learning architectures, and perception systems for embedded platforms, mainly for SoC FPGAs.}

{\includegraphics[width=1in,height=1.25in,clip,keepaspectratio]{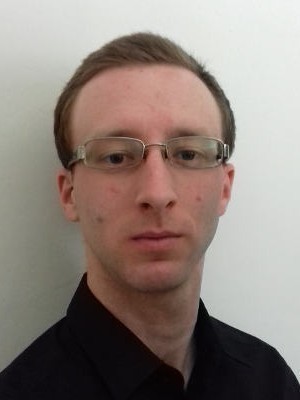}}
{\textbf{Krzysztof Błachut} received the Master of Science
degree in 2019 in Automation and Robotics at AGH
University of Krakow, where is currently pursuing
the Ph.D. degree and working as research assistant.
His professional interests include embedded vision
systems, autonomous vehicles, video surveillance
systems, event cameras and heterogeneous computing platforms such as SoC FPGAs and eGPUs.}

{\includegraphics[width=1in,height=1.25in,clip,keepaspectratio]{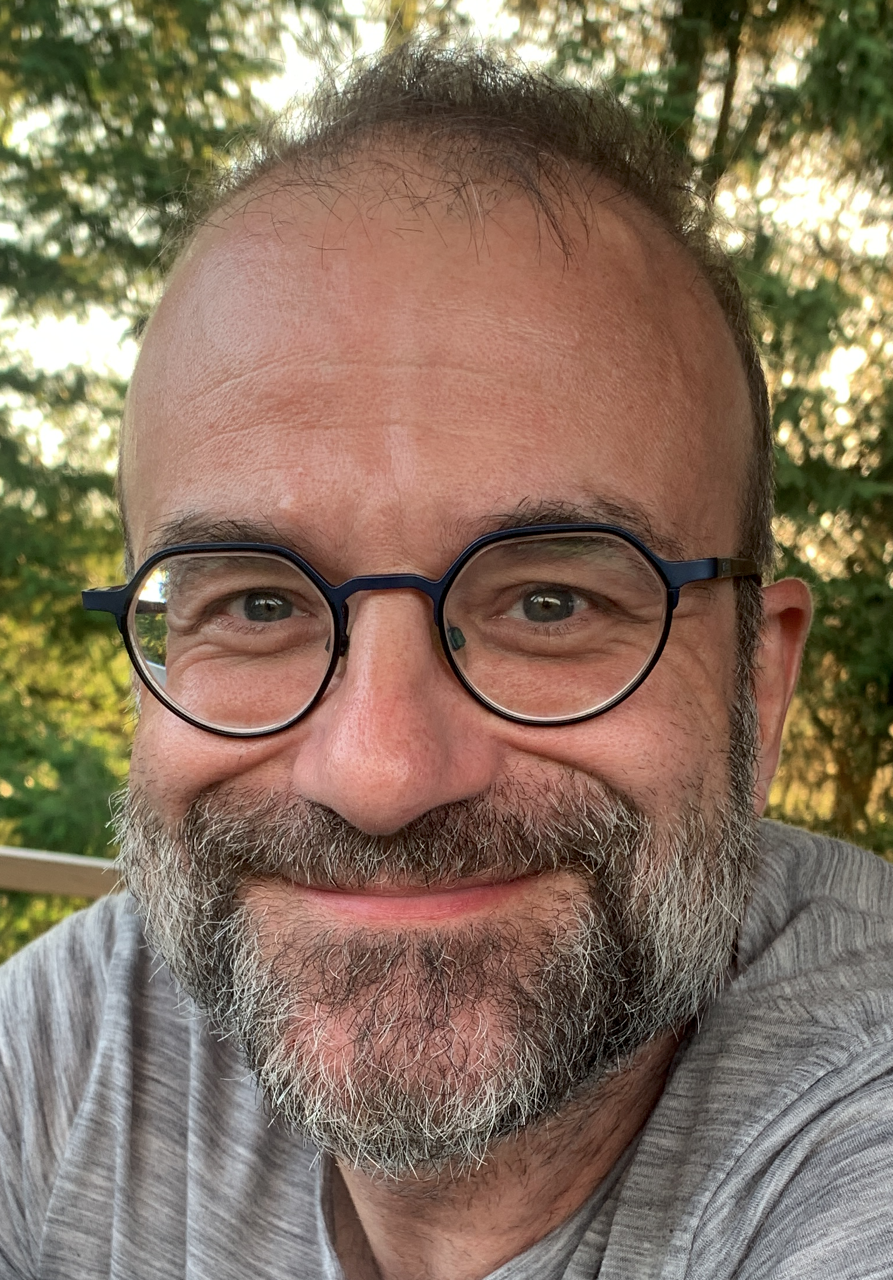}}
{\textbf{Andrea Pinna} received the engineer diploma in
microelectronic from the University of Genoa (Italy),
in 1999, and the Ph.D. degree in electronic system
from the University Pierre and Marie Curie (Paris
– France), in 2003. He was an Associate Professor
from 2004 to 2022 at Sorbonne University (previously Pierre and Marie Curie University) in Paris,
France. Between 2006 and 2011, he worked in the
industrial and semiconductor sector as a Project
Manager for innovative and technology transfer
projects. He joined the Lip6 laboratory at Sorbonne
University in 2011 and has been a full professor since September 2023.
His main research works are designing intelligent medical devices based on
embedded systems to support data analysis and diagnosis. He also explores
reconfigurable and edge-computing architectures and vision system-on-chip
to develop new co-design methods for embedded AI systems.}
\\

{\includegraphics[width=1in,height=1.25in,clip,keepaspectratio]{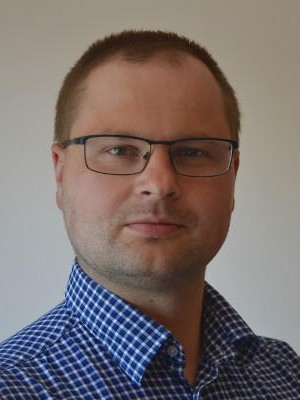}}
{\textbf{Tomasz Kryjak }is an Associate Professor at the Embedded Vision Systems Group, Department of Automatic Control and Robotics, AGH University of Krakow, Poland. His research focuses on embedded perception and control systems implemented in SoC FPGAs, with an emphasis on hardware-aware algorithm design. His application domains span mobile robotics (including drones and autonomous vehicles), Advanced Driver Assistance Systems (ADAS), and intelligent video surveillance.
He works at the intersection of classical frame-based vision, event-based vision, embedded AI, and neuromorphic computing. He is an IEEE Senior Member, a Steering Committee member of the DSD and DASIP conferences, a Technical Program Committee member of the ARC and FPL conferences, and an Associate Editor of the \textit{Microprocessors and Microsystems} journal. He has authored and co-authored over 150 scientific publications.
}

\end{document}


\begin{frontmatter}



\title{Embedded Graph Convolutional Networks for Real-Time Event Data Processing on SoC FPGAs -- Supplementary Material}


\author[agh]{Kamil Jeziorek}
\author[agh]{Piotr Wzorek}
\author[agh]{Krzysztof Blachut}
\author[sorbonne]{Andrea Pinna}
\author[agh]{Tomasz Kryjak}

\affiliation[agh]{organization={
            Embedded Perception and Autonomous Systems Group,\\
            Department of Automatic Control and Robotics, 
            AGH University of Krakow},
            country={Poland}}

\affiliation[sorbonne]{organization={
            Sorbonne Université, 
            CNRS, 
            LIP6, 
            F-75005},
            country={Paris}}



\end{frontmatter}



\section{Model details}

\begin{table}[htp]
\centering
\caption{Detailed architecture configurations of the three EFGCN model variants (Small, Base and Large).}
\resizebox{0.9\columnwidth}{!}{%
\begin{tabular}{@{}lccc@{}}
\toprule
Layer    & Small                 & Base                  & Large                  \\ \midrule
Conv1    & \(1+3 \rightarrow 16\)  & \(1+3 \rightarrow 16\)  & \(1+3 \rightarrow 16\)   \\
3D MaxPool1 & \(4 \times 4 \times 4\)             & \(4 \times 4 \times 4\)             & \(4 \times 4 \times 4\)              \\
Conv2    & \(16+3 \rightarrow 32\) & \(16+3 \rightarrow 32\) & \(16+3 \rightarrow 32\)  \\
Conv3    & \(32+3 \rightarrow 32\) & \(32+3 \rightarrow 32\) & \(32+3 \rightarrow 64\)  \\
3D MaxPool2 & \(2 \times 2 \times 2\)             & \(2 \times 2 \times 2\)              & \(2 \times 2 \times 2\)               \\
Conv4    & \(32+3 \rightarrow 32\) & \(32+3 \rightarrow 64\) & \(64+3 \rightarrow 64\)  \\
Conv5    & \(32+3 \rightarrow 32\) & \(64+3 \rightarrow 64\) & \(64+3 \rightarrow 128\) \\
PoolOut  & \(4 \times 4\)                  & \(4 \times 4\)                  & \(4 \times 4\)                   \\
Linear   & \(16 \times 32 \rightarrow cls\)                & \(16 \times 64 \rightarrow cls\)               & \(16 \times 128 \rightarrow cls\)               \\ \bottomrule
\end{tabular}%
}
\label{table:models_details}
\end{table}

As part of our research, we used three variants of the EFGCN model: Small, Base and Large. Each model consists of five convolutional layers, supplemented by two 3D MaxPool layers, and, after the final convolution, an additional PoolOut layer, which operates solely on the feature vectors of each node, selecting the maximum value from all nodes located within a~defined spatial neighbourhood.

Table \ref{table:models_details} presents detailed information about the models used. In the case of convolutional layers, the $+3$ notation refers to additional spatio-temporal event position features. For the 3D MaxPool layers, the given values define the voxel size in three dimensions. In the PoolOut layer, the operation is performed only in the spatial domain, resulting in just 2 values.

\section{FLOPs calculation}

In this Section, we describe how we calculated the number of FLOPs performed by individual components of the convolution operation.  
Let:  

\begin{itemize}
    \item \(N\) denote the number of nodes in the graph, 
    \item \(E\) denote the number of edges in the graph, 
    \item \(K\) denote the average number of neighbours for an event in the graph,  
    \item \(F_{in}\) and \(F_{out}\) denote the dimensions of the input and output data after the linear layer, respectively.  
\end{itemize}

In this work, we used the PointNetConv layer, which consists of a~single linear MLP layer called for each edge \( e \in E \) in the graph. For a~single edge, it requires multiplications and additions depending on the data dimensions:  
\begin{equation}
    FLOPS_{MLP} = 2 \times F_{in} \times F_{out}.
\end{equation}

Then, for each node in the graph, an aggregation is performed by computing the maximum value from all of its neighbours:  

\begin{equation}
    FLOPS_{Aggr} = F_{out} \times K.
\end{equation}

Finally, feature vector of node is updated, which requires one operation per data dimension:  

\begin{equation}
    FLOPS_{Updt} = F_{out}.
\end{equation}

Taking into account the entire graph and all operations, the total number of FLOPs can be expressed as:
\begin{multline*}
    FLOPS_{Tot} = 2 \times F_{in} \times F_{out} \times E + F_{out} \times K \times E + F_{out} \times E \\
    = E \times (2 \times F_{in} \times F_{out} + F_{out} \times K + F_{out}) \\ 
    = E \times F_{out} \times (2 \times F_{in} + K + 1).
\end{multline*}

The number of FLOPs per event was then determined by taking the total number of operations in the graph and averaging it over the number of events.  

After the MaxPool operation, the values of the number of nodes \(N\), the number of edges \(E\), and the average number of neighbours \(K\) decreased accordingly, which led to a~reduction in the number of FLOPs per event.  

\section{Additional experiments}

\subsection{Radius distance}

One of the key factors during graph generation is the neighbour search radius. Therefore, we conducted an ablation study to evaluate the effectiveness of our solution when using a~larger radius. Following the example of other studies, where a~radius of 5 was applied, we compared these results with our approach using a~radius of 3. The outcomes of this comparison are summarised in Table \ref{tabel:radius_abl}.

The results clearly demonstrate that using a~larger radius allows for improved performance. Comparing our findings with those reported in other works (Table 6 in the main paper), it can be observed that with an increased neighbour search radius, our Large model achieves higher accuracy than the AEGNN model on both datasets. Additionally, for the N-Cars dataset, our models achieves the highest accuracy among all compared approaches, while for the other datasets, the results approach the best reported values.

This confirms how significant the graph structure is for model performance and suggests that future research should also focus on increasing hardware representation capabilities.

\begin{table}[t!]
\caption{Ablation studies over different radius search for graph generation.}
\resizebox{\columnwidth}{!}{%
\begin{tabular}{@{}lccccc@{}}
\toprule
Model   & Radius & N-Cars & N-Caltech101 & CIFAR10-DVS & MNIST-DVS \\ \midrule
EFGCN-S & 3      & 0.907  & 0.623        & 0.571       & 0.964     \\
EFGCN-B & 3      & 0.920  & 0.631        & 0.588       & 0.978     \\
EFGCN-L & 3      & 0.923  & 0.639        & 0.613       & 0.983     \\ \midrule
EFGCN-S & 5      & 0.929  & 0.665        & 0.623       & 0.979     \\
EFGCN-B & 5      & 0.941  & 0.673        & 0.631       & 0.985     \\
EFGCN-L & 5      & 0.947  & 0.688        & 0.654       & 0.991     \\ \bottomrule
\end{tabular}%
}
\label{tabel:radius_abl}
\end{table}

\section{Sequential multiplication strategy}

\begin{table*}[]
\centering
\caption{Selection of parallel multiplication strategy for each layer of \texttt{TIME\_WINDOW} $= 50$ ms models.}
\resizebox{0.99\textwidth}{!}{%
\begin{tabular}{ccc|ccc|ccc|ccc}
\toprule
  \multicolumn{3}{c}{Model details} &
  \multicolumn{3}{c}{Small} &
  \multicolumn{3}{c}{Base} &
  \multicolumn{3}{c}{Large} \\ 
\midrule
Layer & Graph $SIZE$ & \Delta T [\mu s] & output $dim$ & m & duration [\mu s] & output $dim$ & m & duration [\mu s] & output $dim$ & m & duration [\mu s]
\\ \midrule

\textbf{Conv 2} & 64 & 781.25 & 32 & 8 & 737.28 & 32 & 8 & 737.28 & 32 & 8 & 737.28 \\
\textbf{Conv 3} & 64 & 781.25 & 32 & 8 & 737.28 & 32 & 8 & 737.28 & 64 & 16 & 737.28 \\
\textbf{Conv 4} & 32 & 1562.5 & 32 & 1 & 1474.56 & 64 & 2 & 1474.56 & 64 & 2 & 1474.56 \\
\textbf{Conv 5} & 32 & 1562.5 & 32 & 1 & 1474.56 & 64 & 2 & 1474.56 & 128 & 4 & 1474.56 \\
\bottomrule
\end{tabular}
\label{table:tw50_details}
}
\end{table*}

Tables \ref{table:tw50_details} and \ref{table:tw100_details} present calculations conducted to determine the appropriate matrix multiplication parallelisation strategy for each synchronous layer across all supported models and configurations.
The values in the table were determined with Equations (8)-(10) presented in the paper (Section $4.4$).

For \texttt{TIME\_WINDOW} $= 100$ ms, only the Large model was considered, as a~single parallel multiplier ($m=1$) is sufficient for all variants. In contrast, for \texttt{TIME\_WINDOW} $= 50$ ms, we selected the smallest possible value of $m$ that ensures that the throughput requirements for each layer are still met.

\begin{table}[]
\centering
\caption{Selection of parallel multiplication strategy for each layer of \texttt{TIME\_WINDOW} $= 100$ ms models.}
\resizebox{0.9\columnwidth}{!}{%
\begin{tabular}{@{}cccccc@{}}
\toprule
Layer & Graph $SIZE$ & \Delta T [\mu s] & max output $dim$ & m & duration [\mu s] \\ \midrule
Conv2 & 32 & 3125 & 32 & 1 & 1474.56 \\
Conv3 & 32 & 3125 & 64 & 1 & 2949.12  \\
Conv4 & 16 & 6250 & 64 & 1 & 737.28  \\
Conv5 & 16 & 6250 & 128 & 1 & 1474.56  \\ \bottomrule
\end{tabular}%
}
\label{table:tw100_details}
\end{table}

\section{Detailed resource utilisation}

Tables \ref{table:utilisation50} and \ref{table:utilisation100} provide a~detailed resource utilisation analysis for individual modules in the \ours-B model with \texttt{TIME\_WINDOW} $= 50$ ms and \texttt{TIME\_WINDOW} $= 100$ ms configurations.
In the system, internal memory resources are utilised to store neural network weights (within the \texttt{u\_sync\_conv} modules) as well as feature maps between successive network layers (handled by the \texttt{u\_feature\_mem} modules -- implemented using either UltraRAM or BRAM).

As the graph convolutional layers progress through the model, the feature vector dimensions increase, resulting in wider memory data widths. Simultaneously, subsequent 3D MaxPool layers reduce the size of the processed graph, thus decreasing the required memory depth.




\begin{table}[!t]
\centering
\caption{Resource utilisation for the \ours-B model on ZCU104 platform for \texttt{TIME\_WINDOW} $= 50$ ms.}
\resizebox{0.98\textwidth}{!}{%
\begin{tabular}{@{}lccccc@{}}
\toprule
Module & LUT & FF & BRAM & URAM & DSP
\\ \midrule
GCN accelerator (sum) & 138258 & 27176 & 185 & 12 & 184
\\
GCN accelerator  (usage) & 60\% & 5.9\% & 59.3\% & 12.5\% & 10.7\%
\\ \midrule
\texttt{u\_gen\_graph} & 570 & 403 & 17 & 0 & 0
\\
\texttt{u\_async\_conv1} & 5411 & 1193 & 0 & 0 & 64
\\
\texttt{u\_maxpool1} & 773 & 754 & 0 & 0 & 0
\\
\texttt{u\_feature\_mem1} & 517 & 14 & 49.5 & 0 & 0
\\
\texttt{u\_sync\_conv2} & 20849 & 4451 & 2.5 & 0 & 48
\\
\texttt{u\_feature\_mem2} & 1606 & 14 & 0 & 12 & 0
\\
\texttt{u\_sync\_conv3} & 40741 & 7413 & 4.5 & 0 & 48
\\
\texttt{u\_maxpool2} & 931 & 1020 & 0 & 0 & 0
\\
\texttt{u\_feature\_mem3} & 1068 & 12 & 24 & 0 & 0
\\
\texttt{u\_sync\_conv4} & 8906 & 2893 & 4.5 & 0 & 12
\\
\texttt{u\_feature\_mem4} & 2363 & 12 & 45 & 0 & 0
\\
\texttt{u\_sync\_conv5} & 17347 & 4873 & 8 & 0 & 12
\\
\texttt{u\_maxpool3} & 1563 & 1765 & 0 & 0 & 0
\\
\texttt{u\_feature\_mem5} & 1492 & 6 & 30 & 0 & 0
\\
\texttt{u\_out\_serialise} & 211 & 505 & 0 & 0 & 0
\\ \bottomrule
\end{tabular}%
}
\label{table:utilisation50}
\end{table}

\begin{table}[!t]
\centering
\caption{Resource utilisation for the \ours-B model on ZCU104 platform for \texttt{TIME\_WINDOW} $= 100$ ms.}
\resizebox{0.9\textwidth}{!}{%
\begin{tabular}{@{}lcccc@{}}
\toprule
Module & LUT & FF & BRAM & DSP
\\ \midrule
GCN accelerator (sum) & 51936 & 16210 & 159.5 & 88
\\
GCN accelerator  (usage) & 22.5\% & 3.5\% & 51.1\% & 5\%
\\ \midrule
\texttt{u\_gen\_graph} & 515 & 398 & 5.5 & 0
\\
\texttt{u\_async\_conv1} & 5281 & 847 & 0 & 64
\\
\texttt{u\_maxpool1} & 698 & 750 & 0 & 0
\\
\texttt{u\_feature\_mem1} & 698 & 12 & 13.5 & 0
\\
\texttt{u\_sync\_conv2} & 3812 & 1554 & 2.5 & 6
\\
\texttt{u\_feature\_mem2} & 603 & 12 & 24 & 0
\\
\texttt{u\_sync\_conv3} & 7651 & 2263 & 4 & 6
\\
\texttt{u\_maxpool2} & 926 & 1014 & 0 & 0
\\
\texttt{u\_feature\_mem3} & 1322 & 10 & 24 & 0
\\
\texttt{u\_sync\_conv4} & 7115 & 2830 & 4 & 6
\\
\texttt{u\_feature\_mem4} & 1819 & 10 & 45 & 0
\\
\texttt{u\_sync\_conv5} & 10853 & 4264 & 7 & 6
\\
\texttt{u\_maxpool3} & 1561 & 1764 & 0 & 0
\\
\texttt{u\_feature\_mem5} & 1940 & 6 & 30 & 0
\\
\texttt{u\_out\_serialise} & 131 & 440 & 0 & 0
\\ \bottomrule
\end{tabular}%
}
\label{table:utilisation100}
\end{table}